\documentclass[12pt]{article}

\usepackage{amsmath,amssymb,amsthm}
\usepackage{enumitem}
\usepackage{graphicx}
\usepackage{hyperref}
\usepackage{geometry}
\usepackage{setspace}
\usepackage{array}

\usepackage{natbib}
\usepackage{pdflscape}
\usepackage{tabularx, booktabs, amssymb, pifont}
\usepackage{caption}
\usepackage{float}     % ensure [H] works
\usepackage{caption}   % better caption handling
\usepackage{longtable}
\usepackage{booktabs}
\usepackage{array}
\usepackage{ltablex} 
\usepackage{booktabs}
\usepackage{longtable}
\usepackage{fancyhdr}
\keepXColumns          % keeps column widths consistent across pages
\newcommand{\cmark}{\ding{51}} % ✓
\newcommand{\xmark}{\ding{55}} % ✗

\title{Deep Learning for Short-Term Precipitation Prediction in Four Major Indian Cities:\\
A ConvLSTM Approach with Explainable AI}

\author{Tanmay Ghosh\thanks{National Institute of Advanced Studies, Indian Institute of Science campus, Bengaluru, India.
  \texttt{tanmayghosh10@gmail.com}}
  \and
  Shaurabh Anand\thanks{School of Development, Azim Premji University, Bengaluru, India.
  \texttt{shaurabh.anand@apu.edu.in}}
  \and
  Rakesh Gomaji Nannewar\thanks{National Institute of Advanced Studies, Indian Institute of Science campus, Bengaluru, India.
  \texttt{rakesh.nannewar@nias.res.in}}
  \and
  Nithin Nagaraj\thanks{Complex Systems Programme, National Institute of Advanced Studies, Indian Institute of Science campus, Bengaluru, India.
  \texttt{nithin@nias.res.in}}}

\begin{document}
\maketitle
\begin{abstract}
Deep learning models for precipitation forecasting often function as black boxes, limiting their adoption in real-world weather prediction. To enhance transparency while maintaining accuracy, we developed an interpretable deep learning framework for short-term precipitation prediction in four major Indian cities: Bengaluru, Mumbai, Delhi, and Kolkata—spanning diverse climate zones. We implemented a hybrid Time-Distributed CNN-ConvLSTM (Convolutional Neural Network–Long Short-Term Memory) architecture, trained on multi-decadal ERA5 reanalysis data. The architecture was optimized for each city with a different number of convolutional filters: Bengaluru (32), Mumbai and Delhi (64), and Kolkata (128). The models achieved root mean square error (RMSE) values of 0.21 mm/day (Bengaluru), 0.52 mm/day (Mumbai), 0.48 mm/day (Delhi), and 1.80 mm/day (Kolkata). Through interpretability analysis using permutation importance, Gradient-weighted Class Activation Mapping (Grad-CAM), temporal occlusion, and counterfactual perturbation, we identified distinct patterns in the model's behavior. Our results show that the model relied on city-specific variables, with prediction horizons ranging from one day for Bengaluru to five days for Kolkata. This study demonstrates how explainable AI (xAI) can provide accurate forecasts and transparent insights into precipitation patterns in diverse urban environments.

\end{abstract}

\section{Introduction}

India has been witnessing a steady intensification of extreme precipitation events over the past century, which indicates a clear shift in rainfall behaviour across the country \citep{Guhathakurta2011, SenRoy2004}. Long-term station records show that a majority of stations have experienced an increasing trend in extreme rainfall since 1910 \citep{SenRoy2004}, though this intensification has not been uniform; while heavy rainfall events have decreased over parts of central and northern India, they have increased significantly over peninsular, eastern, and northeastern regions, raising the risk of floods in many areas \citep{Guhathakurta2011, Chaubey2022, Ali2014}. Projections from bias-corrected CMIP6 ensembles further suggest near-term (2021--2040) increases of about 4--10\% in the frequency of extreme rainfall, with stronger intensification expected under high-emission scenarios \citep{Chaubey2023}. The increasing frequency and intensity of such events are now recognized as major drivers of recurrent flooding and hydrometeorological hazards across India \citep{Chuphal2025, Chaubey2022, Chuphal2025, Mondal2015}.

In cities, these impacts are further aggravated by rapid urbanization, inadequate stormwater infrastructure, and the intensification of the monsoon under a changing climate \citep{Goswami2006, Roxy2017, Mukherjee2018, Singh2023, Deopa2024}. Observational analyses reveal a growing prevalence of short-duration, high-intensity rainfall episodes across large parts of the Indian subcontinent, particularly during the south-west monsoon season \citep{Saikranthi2024, Fowler2021}. Consequently, major metropolitan areas such as Bengaluru, Mumbai, Kolkata, and Delhi have experienced severe flooding in recent years \citep{Davis2022, Dhiman2019, Paul2018, Dasgupta2013, Mujumdar2021, Bindal2022}. These four cities span contrasting climatic regimes, from the semi-arid northwest to the humid subtropical east, offering a diverse testbed for evaluating how data-driven models perform across multiple monsoon environments.

Based on the growing urban flood risk and the need for city-scale readiness, accurate short-term precipitation forecasts are critical; yet forecasting remains a major challenge in India \citep{Madolli2022, Shejule2022, Pattnaik2019}. Even high-resolution Numerical Weather Prediction (NWP) systems struggle to represent the localised convective processes that drive intense urban rainfall \citep{Davis2022, Shejule2022, Paul2018, Pattnaik2019}. Uncertainties in initial conditions, sparse and uneven observational inputs, and simplifications in physical parameterisations frequently lead to underestimation of heavy rainfall. While physics-based models remain essential for capturing large-scale circulation, their predictive skill degrades at the finer spatial and temporal scales required for city-level flood preparedness. This calls for complementary data-driven methods that learn rainfall dynamics directly from observations rather than relying solely on explicit physical formulations \citep{Singh2023}.

Among recent advances, the Convolutional Long Short-Term Memory (ConvLSTM) model has gained prominence for spatio-temporal rainfall prediction. The ConvLSTM model integrates convolutional operations with recurrent long short-term memory (LSTM) units, enabling simultaneous learning of spatial features and temporal dependencies within precipitation fields. Initially proposed by \citet{Shi2015}, ConvLSTM demonstrated superior performance over fully connected LSTM and optical-flow-based methods for precipitation nowcasting. Subsequent studies have further established its effectiveness for radar-based nowcasting and daily rainfall forecasting \citep{Wang2023}. Recent work has extended ConvLSTM to the Indian context, where data-driven frameworks have improved prediction accuracy in coastal cities such as Mumbai \citep{Kumar2024b, Chaudhuri2024}. ConvLSTM model capacity to represent spatial and temporal rainfall evolution makes it particularly well suited for complex and highly variable systems such as the Indian summer monsoon.

However, deep learning models, like ConvLSTM, often function as black boxes, offering limited insight into the physical mechanisms or predictors that affect their outputs \citep{Huang2023, GonzalezAbad2023, Mamalakis2020}. This lack of transparency restricts their scientific application and makes them less suitable for sensitive climate related applications \citep{Liu2023, Islam2025, Tang2023}. Recent developments in explainable artificial intelligence (XAI) offer diagnostic tools to address these gaps. Techniques such as gradient-based saliency mapping, permutation-based feature importance, and counterfactual perturbation analysis can help identify which input variables, regions, or time steps influence the model predictions the most \citep{Tang2023, Huang2023, Mamalakis2020}. These approaches improve interpretability and allow an assessment of whether model focus aligns with physically meaningful atmospheric processes, such as low-level moisture convergence, temperature anomalies, or pressure gradients.

This study presents an interpretable deep learning framework for short-term precipitation forecasting across four major Indian cities. The proposed hybrid Time-Distributed Convolutional and ConvLSTM architecture captures both spatial and temporal dependencies in multi-decadal meteorological datasets from the ERA5 reanalysis. A weighted loss function is employed to emphasise extreme precipitation events, while interpretability techniques, including Gradient-weighted Class Activation Mapping (Grad-CAM), permutation-based feature importance, time-step occlusion, and counterfactual analysis, are used to examine the physical consistency of the model attention. Through comparative experiments across Bengaluru, Mumbai, Kolkata, and Delhi, the study evaluates both predictive skill and interpretability to understand how the learned relationships between atmospheric variables and precipitation extremes vary across distinct monsoon regimes.

By combining prediction with interpretability of the model, this work contributes to the growing body of research on explainable climate modeling. It demonstrates how deep learning can enhance rainfall forecasting while providing the explainability framework to derive model interpretation, which is consistent with known meteorological processes. This framework bridges the gap between predictive accuracy and scientific interpretability, thus advancing deep learning applications in meteorology and supporting climate-resilient urban planning.

The remainder of this paper is structured as follows: Section 2 provides the necessary background on deep learning for precipitation prediction and interpretable AI. Section 3 details the study areas and the ERA5 reanalysis data used, while Section 4 outlines the methodology for the hybrid ConvLSTM model and the suite of XAI techniques. Section 5 presents and discusses the results, validating both predictive performance and physical interpretability across the four cities. Finally, Section 6 discusses the study's limitations and suggests future work, and Section 7 concludes by summarizing the key findings.

\section{Background}
Over the past decade, the application of deep learning techniques has given rise to considerable success in precipitation forecasting problems around the world. A prominent direction is short-term precipitation nowcasting, where deep neural networks predict high-resolution rainfall from recent radar or satellite observations. Recurrent neural network architectures, especially the Convolutional Long Short-Term Memory (ConvLSTM), have become a cornerstone for such tasks. ConvLSTM networks were first introduced by \citet{Shi2015} to model the evolution of radar echo maps and demonstrated a significant improvement in nowcast accuracy over traditional optical-flow based methods. By integrating convolutional structures into the LSTM cells, ConvLSTM can capture spatiotemporal dependencies in rainfall data, making it well-suited for predicting the motion and growth/decay of storm cells. Subsequent studies have extended this approach with deeper networks and larger datasets. For example, researchers applied ConvLSTM variants to continental-scale radar data in the United States and achieved skillful 6-hour precipitation nowcasts \citep{Sonderby2020}. Generative adversarial models and transformer-based models have also been explored to address limitations like blur in longer lead-time forecasts. The deep generative model by \citet{Ravuri2021} is notable for producing realistic probabilistic radar forecasts up to 90~minutes ahead, outperforming operational nowcasting systems in 89\% of cases according to a panel of meteorologists. These global advancements highlight the potential of deep learning to provide timely, high-resolution predictions of rainfall, especially for convective storms and rapidly evolving weather systems.

In parallel, there is a growing body of work applying deep learning to Indian rainfall forecasting problems, spanning both urban nowcasting and broader monsoon prediction. India’s monsoon climate presents unique challenges such as high seasonal variability and complex regional drivers, but recent studies suggest data-driven models can capture some of these patterns. ConvLSTM-based models have been tested in various Indian contexts. For instance, \citet{Kumar2024a} developed a ConvLSTM nowcasting model using weather radar data over Bhopal (central India) with a 20-minute update cycle. Their model achieved up to 75\% correlation skill for the first 20-minute rainfall prediction and maintained useful skill (~35\% correlation) at a 100-minute lead time, outperforming a persistence baseline and a ConvGRU alternative \citep{Kumar2024a}. At a broader scale, \citet{Kumar2022} trained a ConvLSTM on daily gridded rainfall data (incorporating rain-gauge observations and satellite estimates) to forecast Indian monsoon rainfall up to 2 days ahead. The model showed reasonable accuracy, with correlation coefficients around 0.67 for one-day leads and 0.42 for two-day leads, and higher skill in heavy-rainfall regions like the Western Ghats \citep{Kumar2022}. Other approaches have combined deep learning with atmospheric predictors: \citet{Khan2020} proposed a hybrid 1D CNN and multi-layer perceptron model to predict daily rainfall 1--5 days in advance at locations across Maharashtra, India. This hybrid model outperformed traditional neural networks and support vector regression benchmarks, although forecast error grew with longer lead times. These studies demonstrate the applicability of deep learning for both nowcasting in cities and short-range regional forecasts in the Indian context. However, they also underscore the importance of tailoring models to regional climate characteristics – for example, capturing the diurnal cycle of convection or the influence of monsoon troughs – and the need for extensive evaluation under different rainfall regimes \citep{Li2022}.

\subsection{Explainable AI for Rainfall Prediction}

While deep learning models have proven highly effective for precipitation prediction, their intricate, non-linear nature often renders them ``black boxes,'' (\citet{mcgovern2019making} hindering trust and adoption in operational meteorology \cite{Yang2024}. Complex models like ConvLSTMs or generative networks can be difficult to interpret, making it challenging to understand why specific rainfall predictions are made. This opaqueness presents particular challenges for operational forecasting, where meteorologists and disaster managers require interpretable model outputs to make critical decisions.

To address these limitations, recent research has developed two complementary approaches: explainable AI (XAI) techniques and hybrid modeling strategies. A crucial distinction lies between interpretability (designing models whose internal structures align with physical mechanisms) and explainability (using post-hoc tools to attribute specific forecasts to inputs or regions) \cite{Mamalakis2020}. Post-hoc explanation tools such as Shapley Additive exPlanations (SHAP) and Grad-CAM have been applied to highlight which input features (e.g., particular radar cells or atmospheric variables) most influence neural network predictions \cite{Yang2024}. Such methods can reveal, for instance, that a model has learned to focus on upstream convective cells when predicting rainfall for a city an hour later. Simultaneously, researchers have pursued strategies to integrate physical knowledge directly into model architecture. Examples include using attention mechanisms to align model focus with known weather patterns, or developing neural networks that respect conservation laws to enhance physical interpretability \cite{Reichstein2019}.

Interpretability is often engineered through architectural inductive biases. For instance, \citeauthor{Yang2022} developed a spatio-temporal graph-guided ConvLSTM that explicitly encodes dynamic relationships between tropical cyclone features and precipitation regions, significantly improving heavy rainfall prediction by propagating information along physically plausible pathways \cite{Yang2022}. For longer-range forecasts, \citeauthor{Jiang2024} proposed a CNN--Transformer--ConvLSTM hybrid for post-processing numerical weather prediction data. This architecture leverages local convolutions for fine-scale patterns, attention mechanisms for synoptic-scale context, and recurrent layers for temporal memory, employing a quantile-weighted loss to emphasize extreme events \cite{Jiang2024}. Furthermore, \citeauthor{Liu2024} demonstrated that integrating dedicated input channels for meteorological drivers, such as tropical cyclone tracks, creates an inspectable signal path, enabling auditors to trace how track errors propagate into rainfall and subsequent flood forecasts \cite{Liu2024}.

In precipitation forecasting specifically, studies have explored hybrid models that combine data-driven and physics-based approaches, such as using machine learning to post-process NWP outputs or creating ensembles of neural networks and numerical models to quantify uncertainty \cite{Ravuri2021, Yang2024}. These approaches seek to leverage the strengths of both paradigms: the interpretability and theoretical foundation of physics-based models combined with the pattern-recognition capabilities of deep learning.

Post-hoc techniques provide critical verification for individual forecasts. Gradient-weighted Class Activation Mapping (Grad-CAM) generates spatial saliency maps, offering a visual check of a model's focus. In basin-scale flood forecasting, an explainable CNN using Grad-CAM consistently highlighted precipitation and inflow patterns during peak flows, confirming its reliance on physically sensible evidence \cite{Xiang2024}. Occlusion sensitivity analysis quantifies the dependence on specific inputs by measuring performance degradation when features are masked. This method has been used to quantify the critical importance of track data in short-term cyclone nowcasting and to audit error propagation through the forecasting chain \cite{Liu2024}. For models with heterogeneous predictors, SHAP provides stable variable importance rankings. \citeauthor{Wu2023} used SHAP in a runoff ensemble to distinguish long-term trend drivers from short-term volatility, enhancing transparency without sacrificing skill \cite{Wu2023}.

Three key principles for interpretable AI in meteorology emerge from the literature \citep{Liu2024, Yang2022, Huang2024, Xiang2024, Jiang2024}. First, model architectures should be designed to reflect physical relationships in atmospheric processes. Second, interpretability frameworks should provide explanations at multiple scales, from global summaries of model behavior to detailed analyses of individual prediction events. Third, model development must emphasize accurate forecasting of extreme weather events where operational risks are greatest.

Building on these principles, this study implements an integrated interpretability framework specifically tuned to each city's meteorological characteristics. We combine four complementary techniques: permutation importance to identify critical predictive variables, counterfactual perturbation to assess prediction sensitivity, temporal occlusion to determine memory dependencies, and Grad-CAM to visualize spatial attention patterns. This comprehensive approach enables detailed examination of model behavior across different cities and rainfall regimes, providing both broad understanding of operational logic and specific insights into individual forecasts.

\setlength{\LTpre}{1pt}
\setlength{\LTpost}{1pt}

\begin{longtable}{%
  p{0.16\textwidth}%
  p{0.18\textwidth}%
  p{0.18\textwidth}%
  p{0.12\textwidth}%
  p{0.30\textwidth}%
}
\caption{Representative deep learning studies on precipitation forecasting, illustrating regions, methods, lead times, and performance.}
\label{tab:dl_studies}\\
\toprule
\textbf{Study (Year)} &
\textbf{Region / City} &
\textbf{Method} &
\textbf{Lead Time} &
\textbf{Performance Metrics} \\
\midrule
\endfirsthead

\caption[]{Representative deep learning studies on precipitation forecasting (continued).}\\
\toprule
\textbf{Study (Year)} &
\textbf{Region / City} &
\textbf{Method} &
\textbf{Lead Time} &
\textbf{Performance Metrics} \\
\midrule
\endhead

\midrule
\multicolumn{5}{r}{\emph{Continued on next page}}\\
\bottomrule
\endfoot

\bottomrule
\endlastfoot

Shi et al.\ (2015) &
Hong Kong (local nowcast) &
ConvLSTM (radar echo) &
0--1\,hour (6--10\,min step) &
Outperformed optical-flow baseline; higher CSI and correlation than ROVER for heavy rain nowcasts. \\
\addlinespace

Sønderby et al.\ (2020) &
Continental USA &
MetNet (CNN + Attention) &
0--8\,hours (2\,min step) &
Improved threat scores vs.\ NOAA's HRRR up to 7--8\,h; $\sim$10--20\% lower MSE than NWP at short lead times. \\
\addlinespace

Ravuri et al.\ (2021) &
UK (national nowcasting) &
DGMR generative model &
5--90\,min &
Ranked 1st in 89\% of cases in expert evaluation; no blurring at 90\,min lead. \\
\addlinespace

Khan \& Maity (2020) &
Maharashtra, India (12 sites) &
Hybrid 1D-CNN + MLP &
1--5\,days (daily) &
MAE $\approx$10--15\% of mean rainfall; outperformed MLP and SVR; skill decreases with lead time. \\
\addlinespace

Kumar et al.\ (2022) &
All-India (monsoon regions) &
ConvLSTM (daily grids) &
1--2\,days (daily) &
CC = 0.67 (1-day), 0.42 (2-day); RMSE $\sim$10--20\,mm; higher skill over Western Ghats. \\
\addlinespace

Kumar et al.\ (2024a) &
Bhopal, India (urban nowcast) &
ConvLSTM vs.\ ConvGRU &
0--100\,min (20\,min step) &
75\% correlation at 20\,min, $\sim$35\% at 100\,min; higher CSI and lower RMSE than persistence. \\
\addlinespace

Xiang et al.\ (2024)  &
Lushui Basin, China  &
Explainable CNN (ECNN) with Grad-CAM  &
3--12 hours (80 time-steps, 240 hours) &
Outperformed LSTM, CNN, and LSTM-CNN models; NSE: 0.96--0.81 (test period); RMSE: 32.25--64.37 m³/s; RE: $\pm$2.88\%; interpretable via Grad-CAM. \\ 
\addlinespace

Huang et al.\ (2024) &
Beijiang River Basin, China &
Coupled CNN-LSTM with SHAP &
1--25 hours &
Outperformed CNN and LSTM; NSE: 0.838 (25 h forecast); R²: 0.847--0.972; interpretable via SHAP; downstream runoff identified as most influential input. \\
\addlinespace

Jadhav et al.\ (2024) &
India (general) &
LSTM with SHAP \& LIME &
Not specified (uses 15-day input window) &
LSTM Accuracy: 94.21\%, MSE: 9.7, RMSE: 3.12; SHAP \& LIME identified precipitation hours and max temperature as most influential features. \\
\addlinespace

Yang et al.\ (2022) &
Southeast China &
Spatio-temporal graph-guided ConvLSTM (TC precipitation) &
0--3 hours (30\,min step) &
F1-score improvements of 2.74\%, 4.24\%, 6.29\%, 5.78\% at 5, 10, 20, 30 mm/h thresholds compared to ConvLSTM baseline. \\
\addlinespace

Jiang et al.\ (2024) &
Southeastern China &
TransLSTMUNet (TIGGE post-processing) &
24--120 hours (24-hour step) &
Enhanced ACC by 12.14\%; TS improvements of 8.30\%, 9.77\%, 31.60\%, 51.25\% for 0.1, 10.0, 25.0, 50.0 mm thresholds respectively. \\
\addlinespace

Pegion et al.\ (2022) &
Southeastern U.S. &
CNN with gridded fields &
Daily (simultaneous predictors) & 
CNN: 70\% accuracy (winter), 60\% (summer); reliable probability forecasts; LR with indices: $\sim$50\% accuracy (not reliable). \\
\addlinespace

Yin et al.\ (2024) &
Jinsha River basin, China &
ZR-ANN, ZR-SVR (radar reflectivity) &
1 hour (10 inputs at 6-min intervals) &
R² scores: ZR-ANN (0.8563), ZR-SVR (0.8557), revised Z-R (0.7840); ZR-ANN outperformed traditional Z-R by 9.2\% in R². \\
\addlinespace

Liu et al.\ (2024) &
Lishimen Reservoir, China &
TCRainNet (ConvLSTM with tropical cyclone track fusion) &
0--6 hours (hourly) &
POD~$>$~0.27, CSI~$>$~0.20, MAE~$<$~2.6~mm (46\% lower than ECMWF-HRES), NSE~$>$~0.70 for flood forecasts with a +4~h lead time. \\
\addlinespace

Kumar et al.\ (2020) &
Global (satellite-based, 150$\times$150 km grids) &
Convcast (ConvLSTM with AutoML tuning, IMERG satellite data) &
30--150 minutes (30-min steps, iterative) &
30-min: Accuracy=0.93, RMSE=0.805 mm/h, HSS=0.801; 150-min: Accuracy=0.87, RMSE=1.389 mm/h, HSS=0.623. Outperformed optical flow methods (Sparse, Dense, DenseROT, SparseSD). \\
\addlinespace

Miao et al.\ (2019) &
Xiangjiang River Basin, China &
ConvLSTM (CNN + LSTM) &
1--15 days &
CC: 0.58 (Day 1) to 0.21 (Day 15); Improved hydrological simulation NSE from 0.06 (raw ERA-Interim) to 0.64 (corrected precipitation). Outperformed QM, SVM, and CNN. \\
\end{longtable}

\section{Data and Study Area}

This study examines short-term precipitation dynamics across four major Indian metropolitan regions, Delhi, Mumbai, Kolkata, and Bengaluru, each representing a distinct climatic regime. Delhi lies in the semi-arid continental zone of northern India, characterized by large diurnal temperature variations and limited monsoon rainfall \citep{IMD2013, Chevuturi2015}. Mumbai, located on the western coast, experiences a tropical monsoon climate with over 2400 mm of annual rainfall, primarily during June–September \citet{Mann2023}. Kolkata, in eastern India, has a humid subtropical climate influenced by the Bay of Bengal branch of the southwest monsoon, while Bengaluru, situated on the southern Deccan Plateau, experiences a transitional semi-arid climate with moderate seasonal precipitation\citet{De2022, Chaudhary2017}. Together, these cities capture the major gradients of monsoon behavior across India, offering a diverse testbed for evaluating the robustness of deep learning models under different climatic and geographic conditions.

\begin{figure}[htbp]
    \centering
    \includegraphics[width=\textwidth,
                     height=0.8\textheight,
                     keepaspectratio]{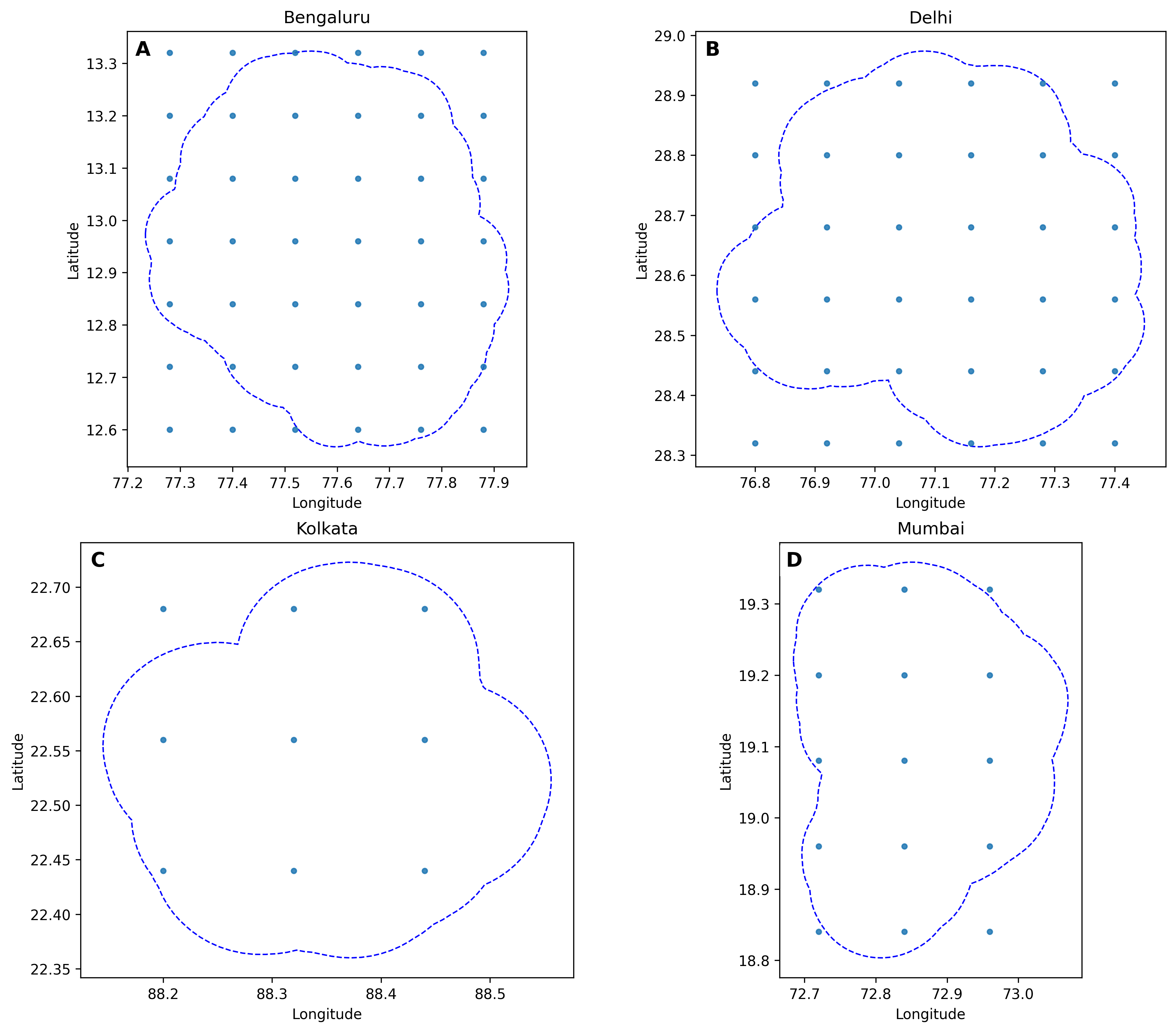}
    \caption{Study area and data points }
    \label{fig:study-area}
\end{figure}

\noindent
This study uses data from the European Centre for Medium-Range Weather Forecasts (ECMWF) ERA5 reanalysis, which provides a spatio-temporally coherent set of key meteorological variables\citet{Rani2021}. Twenty-two years (1998--2020) of data corresponding to the southwest monsoon season (June to September) were extracted for analysis\citet{Rani2021}. The native six-hourly temporal resolution data were aggregated to daily resolution for all subsequent processing. The spatial resolution is 0.25° × 0.25°, with the number of latitude--longitude grid points varying by city as shown in Table~\ref{tab:data_points}. Data from all grid points within each city's domain (10 Kilometers from the city boudary) were obtained and processed.

\begin{table}[htbp]
  \centering
  \small
  \setlength{\tabcolsep}{5pt}
  \renewcommand{\arraystretch}{0.9}
  \caption{Number of data points by city}
  \label{tab:data_points}
  \begin{tabular}{@{}lr@{}}
    \toprule
    \textbf{City} & \textbf{Number of data points} \\
    \midrule
    Mumbai     & 15 \\
    Kolkata    & 9 \\
    Delhi      & 36 \\
    Bangalore  & 42 \\
    \bottomrule
  \end{tabular}
\end{table}
\vspace{-2mm}

We used fourteen meteorological variables as potential predictors, with the total precipitation rate designated as the dependent variable. These predictors represent key atmospheric processes including thermodynamic quantities (surface temperature, relative humidity, total cloud cover), dynamic circulation patterns (zonal and meridional wind components, surface pressure), convective indicators (convective rain rate), and synoptic-scale features (mean sea level pressure, large-scale rain rate). The complete variable list is provided in Table 3.

Data preprocessing included temporal alignment across all variables, handling of missing values, and scaling of predictors to a standardised range. For the target variable, data from all grid points within each city's domain were aggregated to generate area-averaged daily time series. This spatial aggregation reduces noise from localized convective events while maintaining representative regional precipitation patterns. All variables were spatially aligned and normalized to ensure comparability across time and space before being input to the model.

The processed dataset was partitioned chronologically into training, validation, and testing subsets. This temporal splitting preserves the sequence of meteorological events and prevents information leakage from future data. The multi-city, multi-decadal dataset thus provides a comprehensive empirical foundation for developing and validating the proposed interpretable deep learning framework.

\begin{landscape}
\begin{table}[htbp]
\centering
\scriptsize
\caption{Meteorological variables and their physical processes used for model development across four Indian cities}
\label{tab:variables}
\setlength{\tabcolsep}{2.0pt}
\renewcommand{\arraystretch}{1.25}

\begin{tabularx}{1.10\textwidth}{
    >{\raggedright\arraybackslash}p{3.1cm}   % Variable
    >{\raggedright\arraybackslash}p{4.2cm}   % Description
    c                                        % Unit
    @{\hspace{5pt}}>{\centering\arraybackslash}p{0.75cm} % BLR
    @{\hspace{10pt}}>{\centering\arraybackslash}p{0.75cm} % MUM
    @{\hspace{10pt}}>{\centering\arraybackslash}p{0.75cm} % KOL
    @{\hspace{5pt}}>{\centering\arraybackslash}p{0.75cm}  % DEL
    >{\raggedright\arraybackslash}p{3.0cm}   % Physical Process
    >{\centering\arraybackslash}p{1.9cm}     % Resolution
}
\toprule
\textbf{Variable} & \textbf{Description} & \textbf{Unit} &
\textbf{BLR} & \textbf{MUM} & \textbf{KOL} & \textbf{DEL} &
\textbf{Physical Process} & \textbf{Resolution} \\
\midrule
convective\_rain\_rate & Convective precipitation rate & mm/day & \cmark & \cmark & \cmark & \cmark & Convective & 0.25°×0.25°, 6-hourly \\
evaporation\_rate & Evaporation rate & mm/day & \cmark & \cmark & \cmark & \cmark & Surface–Atmosphere Interaction & 0.25°×0.25°, 6-hourly \\
large\_scale\_rain\_rate & Large-scale precipitation rate & mm/day & \cmark & \cmark & \cmark & \cmark & Synoptic & 0.25°×0.25°, 6-hourly \\
relative\_humidity & Relative humidity & \% & \cmark & \cmark & \cmark & \cmark & Thermodynamic & 0.25°×0.25°, 6-hourly \\
surface\_pressure & Surface pressure & Pa & \cmark & \cmark & \cmark & \cmark & Dynamic & 0.25°×0.25°, 6-hourly \\
total\_cloud\_cover & Total cloud cover & fraction & \cmark & \cmark & \cmark & \cmark & Thermodynamic & 0.25°×0.25°, 6-hourly \\
total\_precipitation & Total precipitation (target) & mm/day & \cmark & \cmark & \cmark & \cmark & Target Variable & 0.25°×0.25°, 6-hourly \\
wind\_u & Zonal wind component & m/s & \cmark & \cmark & \cmark & \cmark & Dynamic & 0.25°×0.25°, 6-hourly \\
wind\_v & Meridional wind component & m/s & \cmark & \cmark & \cmark & \cmark & Dynamic & 0.25°×0.25°, 6-hourly \\
mean\_sea\_level\_pressure & Mean sea-level pressure & Pa & \cmark & \cmark & \cmark & \xmark & Synoptic & 0.25°×0.25°, 6-hourly \\
temp\_surface & Surface temperature & K & \cmark & \cmark & \xmark & \cmark & Thermodynamic & 0.25°×0.25°, 6-hourly \\
surface\_temp\_skin & Skin temperature & K & \xmark & \cmark & \cmark & \cmark & Thermodynamic & 0.25°×0.25°, 6-hourly \\
wind\_speed & Wind speed & m/s & \cmark & \xmark & \xmark & \xmark & Dynamic & 0.25°×0.25°, 6-hourly \\
wind\_speed\_10m & Wind speed at 10 m & m/s & \xmark & \cmark & \cmark & \cmark & Dynamic & 0.25°×0.25°, 6-hourly \\
wind\_gust & Wind gust & m/s & \xmark & \cmark & \xmark & \xmark & Dynamic & 0.25°×0.25°, 6-hourly \\
\bottomrule
\end{tabularx}

\vspace{2mm}
\noindent\makebox[\linewidth][c]{%
\parbox{0.88\paperwidth}{%
\footnotesize
\raggedright
\textit{Note:} BLR = Bengaluru, MUM = Mumbai, KOL = Kolkata, DEL = Delhi. 
\cmark{} indicates variable used, \xmark{} indicates variable not used. 
Variables are grouped by dominant physical processes: Thermodynamic (heat/moisture), Dynamic (motion), Convective (localized storms), Synoptic (large-scale systems), and Surface–Atmosphere Interaction (land–air coupling). 
All data are from ERA5 reanalysis (1998–2020) for the monsoon season (June–September).
}}
\end{table}
\end{landscape}

% -- First figure --
\begin{figure}[htbp]
    \centering
    \includegraphics[width=\textwidth,
                     height=1.1\textheight,
                     keepaspectratio]{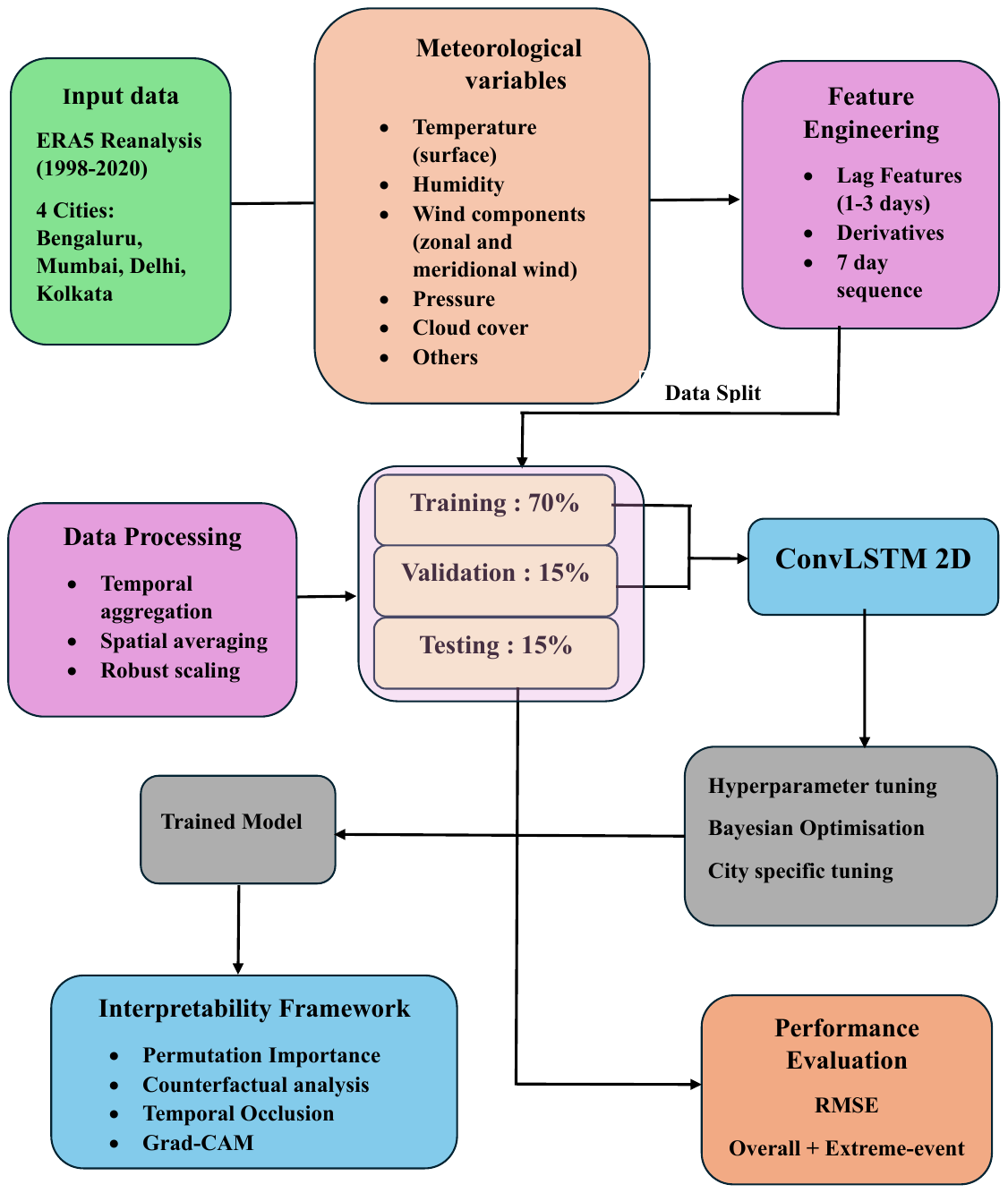}
    \caption{Flowchart used in the present study}
    \label{fig:study_flowchart}
\end{figure}

\begin{figure}[htbp]
    \centering
    \includegraphics[width=1\textwidth]{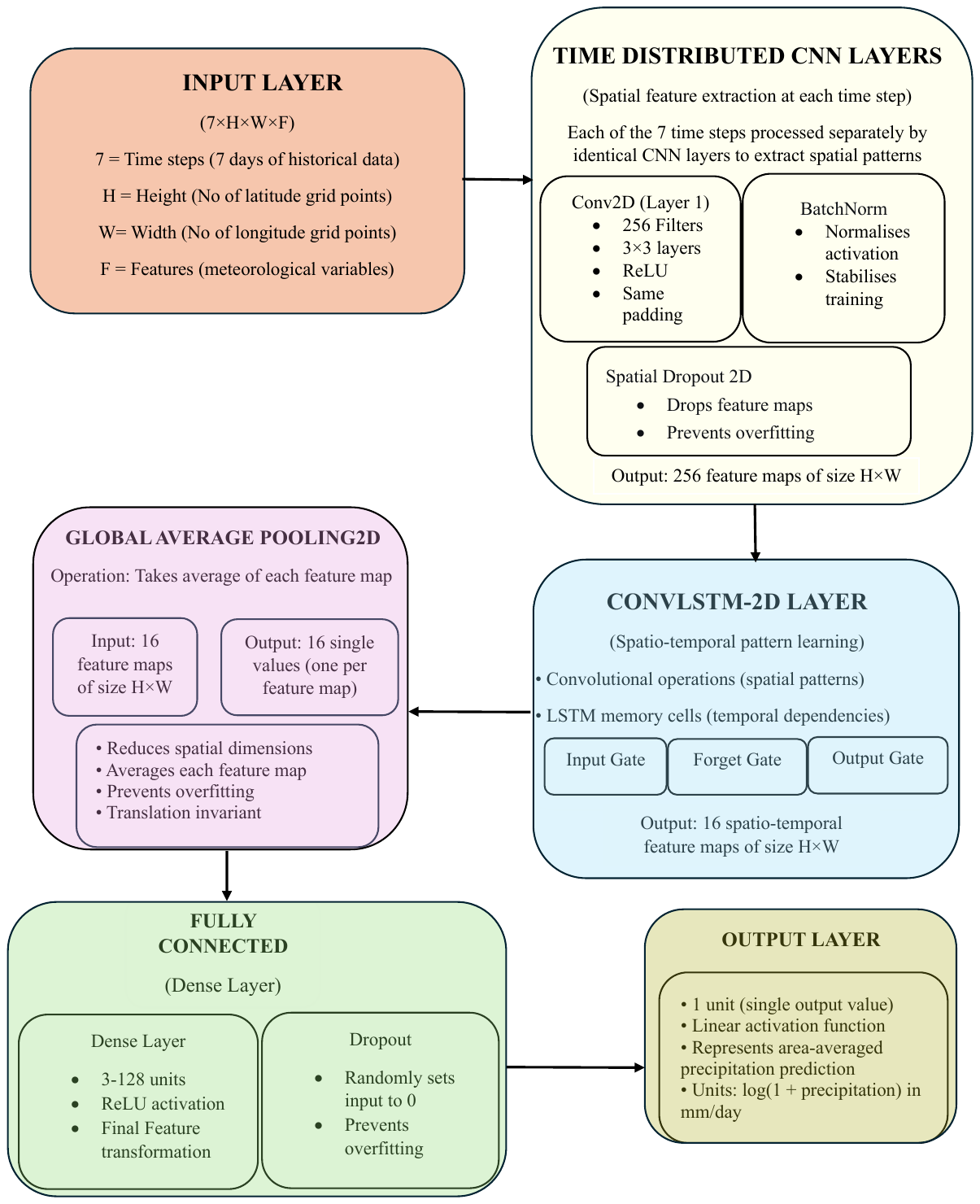}
    \caption{Schematic illustration of the model used in the study}
    \label{fig:model_architecture}
\end{figure}

\section{Methods}

\subsection{Model Framework Overview}

We developed a hybrid deep learning framework that integrates TimeDistributed Convolutional Neural Networks (CNNs) with Convolutional Long Short-Term Memory (ConvLSTM) layers to forecast daily precipitation over major Indian cities. The model captures both spatial and temporal dependencies in reanalysis-based meteorological variables, with architecture choices motivated by precipitation physics: local triggering conditions and multi-day system evolution.

Each input sample is represented as a four-dimensional tensor $X \in \mathbb{R}^{T \times H \times W \times F}$ with dimensions corresponding to time steps, latitude, longitude, and predictor variables. 

\textbf{Temporal Context through Feature Engineering:} To enhance the model's ability to capture temporal evolution, we incorporate lagged versions of meteorological variables from the past 1-3 days. This provides the model with explicit information about recent atmospheric conditions, which is crucial because precipitation often results from processes that unfold over multiple days, such as the slow accumulation of moisture or the movement of large-scale weather systems. These lagged features complement the ConvLSTM's inherent ability to learn temporal patterns.

\textbf{Spatial Processing Strategy:} A key methodological innovation is our dual approach to spatial information. While the model processes full spatial grids ($H \times W$) as input to learn local atmospheric patterns, it predicts area-averaged precipitation as the target variable. This design aligns with urban flood forecasting needs while enabling clear spatial interpretability. This approach enables the model to:
\begin{itemize}
    \item Learn from spatial gradients and local patterns across the urban region
    \item Identify which geographical subregions contribute most to area-averaged rainfall through interpretability methods
    \item Maintain computational efficiency while capturing spatial complexity
\end{itemize}

The spatial processing occurs through convolutional filters embedded within a TimeDistributed wrapper, ensuring that spatial feature extraction occurs independently for each time step while maintaining temporal coherence. The extracted feature maps are subsequently processed by a ConvLSTM layer, which extends the traditional LSTM structure by introducing convolutional operations within its recurrent gates. This design allows the model to retain temporal memory while preserving local spatial correlations that are critical for representing the evolution of convective systems and monsoon dynamics.

Following the ConvLSTM layer, a global average pooling operation aggregates the learned features, which are then passed through fully connected dense layers to produce an area-averaged daily precipitation estimate. The model is trained end-to-end, enabling joint optimization of spatial and temporal representations.This architecture provides the foundation for the interpretability analyses that reveal city-specific precipitation drivers in Section 5.

\subsection{Data Processing and Spatial-Temporal Sequencing}

The input data processing pipeline transforms raw meteorological variables into spatiotemporal sequences suitable for ConvLSTM processing:

\textbf{Spatial Grid Construction:} For each city, data from 64 grid points are organized into 2D spatial maps representing the geographical distribution of atmospheric variables. The spatial dimensions ($H \times W$) preserve the relative positions and neighborhood relationships between grid points.

\textbf{Temporal Feature Engineering:} To capture the temporal evolution of atmospheric conditions, we created lagged versions of all meteorological variables. For each of the twelve core variables, we generated lag features for 1, 2, and 3 previous days. This temporal expansion allowed the model to access recent historical conditions, effectively providing a multi-day context for each prediction. The lag feature creation followed the formula:

\[
X_{t-lag} = \text{shift}(X_t, \text{lag})
\]

where $X_t$ represents each meteorological variable at time $t$, and the shift operation was applied within each geographical grid point's time series to maintain spatial coherence.

Additionally, for precipitation specifically, we derived rate-of-change features by computing differences between consecutive lag days:

\[
\Delta P_{lag} = P_{t-lag} - P_{t-(lag-1)}
\]

where $P$ represents precipitation. These derivative features captured the acceleration or deceleration of precipitation patterns, providing the model with trend information beyond absolute values.

\textbf{Temporal Sequencing:} Input sequences of length $T=7$ days are constructed, with each time step containing the full spatial grid of meteorological variables. This allows the model to learn both the spatial patterns and their temporal evolution.

\textbf{Target Formulation:} The prediction target is the spatial average of precipitation across all grid points for the subsequent day. This area-averaged approach provides several advantages:
\begin{itemize}
    \item Reduces noise from localized, stochastic precipitation events
    \item Aligns with practical forecasting needs for urban-scale water management
    \item Enables meaningful evaluation against observational data
    \item Facilitates interpretability by revealing which spatial regions drive area-wide predictions
\end{itemize}

\textbf{Normalization and Transformation:} All input features undergo robust scaling to mitigate outlier effects, while precipitation values are log-transformed as $\log(1 + y)$ to stabilize variance across the wide range of rainfall intensities.

\subsection{Mathematical Formulation}

Let $X_t \in \mathbb{R}^{H \times W \times C}$ denote the multivariate meteorological variable at time step $t$, where $H$ and $W$ represent spatial dimensions, and $C$ is the number of predictor variables. The model receives an input sequence $\{X_{t-T+1}, \ldots, X_t\}$ of length $T$ and predicts a scalar precipitation value $\hat{y}_{t+1}$ for the subsequent day.

Each TimeDistributed CNN layer performs convolutional operations on $X_t$ to extract spatial feature maps:
\[
F_t = \text{ReLU}(W_c \ast X_t + b_c)
\]
where $W_c$ and $b_c$ are the convolutional kernel and bias, respectively, and $\ast$ represents the convolution operator.

The resulting feature sequence $\{F_t\}$ is processed by a ConvLSTM layer through convolutional gating mechanisms:
\begin{align*}
i_t &= \sigma(W_{xi} \ast F_t + W_{hi} \ast H_{t-1} + b_i) \\
f_t &= \sigma(W_{xf} \ast F_t + W_{hf} \ast H_{t-1} + b_f) \\
o_t &= \sigma(W_{xo} \ast F_t + W_{ho} \ast H_{t-1} + b_o) \\
C_t &= f_t \odot C_{t-1} + i_t \odot \tanh(W_{xc} \ast F_t + W_{hc} \ast H_{t-1} + b_c) \\
H_t &= o_t \odot \tanh(C_t)
\end{align*}
where $i_t$, $f_t$, and $o_t$ denote input, forget, and output gates; and $\odot$ indicates element-wise multiplication.

The final prediction is obtained through global average pooling and dense layers:
\[
\hat{y}_{t+1} = W_d \cdot \mathcal{G}(H_T) + b_d
\]

To address class imbalance in extreme rainfall events, the model minimizes a weighted mean squared error loss:
\[
\mathcal{L} = \frac{1}{N} \sum_{i=1}^{N} w_i (y_i - \hat{y}_i)^2, \quad w_i = \begin{cases} 
\alpha & \text{if } y_i \geq \tau \\
1 & \text{otherwise}
\end{cases}
\]
where $\tau$ is the 90th percentile rainfall threshold and $\alpha > 1$ emphasizes extreme events.

\subsection{Hyperparameter Optimization Strategy}

A systematic Bayesian Optimization approach was employed to automatically tailor model architecture to each city's unique precipitation characteristics. The search space encompassed convolutional capacity (32-128 filters), temporal modeling complexity (16-64 ConvLSTM filters), regularization strength (dropout rates 0.0-0.5), learning rates ($10^{-3}$ to $10^{-4}$), and kernel sizes (3×3 or 5×5). For each cross-validation fold, 20 optimization trials were conducted using validation loss as the objective function, with early stopping to prevent overfitting.

\subsection{Model Training and Validation}

We performed Model training using a temporally ordered sequence to maintain the causal structure inherent in climatic data. For each city, we divided the dataset chronologically into training, validation, and test subsets (70:15:15). This split prevented information leakage from future data. The model underwent 30 training epochs using the Adam optimizer, with the application of early stopping (patience=3) and learning rate reduction on plateau (factor=0.5). To ensure robust generalization, we performed three-fold time-series cross-validation, and used mean out-of-sample RMSE as the primary performance metric.

\subsection{Interpretability Framework}

To bridge the gap between predictive performance and physical understanding, we implemented a comprehensive suite of explainable AI techniques:

\textbf{Spatial Attention via Grad-CAM} identifies geographical regions most influential to predictions by computing gradient-weighted activations from ConvLSTM feature maps\cite{selvaraju2020gradcam}.

\textbf{Permutation Feature Importance} quantifies variable relevance through RMSE increase when individual features are randomly shuffled.

\textbf{Temporal Occlusion Analysis} evaluates time-step importance by measuring prediction degradation when historical data is masked.

\textbf{Counterfactual Explanations} determine minimal feature perturbations required to alter predictions, providing causal sensitivity analysis.

\section{Results and Discussion}

\subsection{Overview of Model Performance Across Urban Regimes}

The ConvLSTM model demonstrated spatial and temporal learning in India's diverse urban rainfall regimes. The performance of the model, in terms of the predictability of precipitation, varied across study regions. (Table~\ref{tab:performance}). 

For the overall rainfall predictions, the Root Mean Square Error (RMSE) was the lowest (0.2088 mm day$^{-1}$) in the case of Bangalore.  Delhi and Mumbai showed relatively higher RMSE (0.4842 mm day$^{-1}$  and 0.5184 mm day$^{-1}$, respectively). The highest RMSE (1.7957 mm day$^{-1}$) was recorded in the case of  Kolkata. The predictability in the case of extreme precipitation differed across cities under study. The RMSE for extreme precipitation was almost double of overall RMSE in the case of  Bengaluru (0.4700  mm day$^{-1}$) , Mumbai (1.2610  mm day$^{-1}$), and Delhi (0.8622  mm day$^{-1}$). The prediction for extreme precipitation was better (RMSE 1.3121  mm day$^{-1}$) than overall precipitation in the case of Kolkata.

\begin{table}[h!]
\centering
\caption{Predictive performance across urban climatic regimes}
\begin{tabular}{lcc}
\hline
\textbf{City} & \textbf{RMSE (mm day$^{-1}$)} & \textbf{Extreme-event RMSE (mm day$^{-1}$)} \\
\hline
Bengaluru & 0.2088 & 0.4700 \\
Mumbai & 0.5184 & 1.2610 \\
Delhi & 0.4842 & 0.8622 \\
Kolkata & 1.7957 & 1.3121 \\
\hline
\end{tabular}
\label{tab:performance}
\end{table}

\subsection{Hyperparameter Optimization}

The Bayesian optimisation process showed distinct architectural preferences that align physically with each city's precipitation characteristics (Table~\ref{tab:hyperparams}). This automated tuning demonstrated the model's ability to adapt to diverse climate regimes.

\begin{table}[h!]
\centering
\caption{Optimal model configurations across urban climates}
\begin{tabular}{lcccc}
\hline
\textbf{City} & \textbf{Conv Filters} & \textbf{ConvLSTM Filters} & \textbf{Dropout Rate} & \textbf{Learning Rate} \\
\hline
Bengaluru & 32 & 16 & 0.4 & 1e-4 \\
Mumbai & 64 & 32 & 0.2 & 1e-3 \\
Delhi & 64 & 32 & 0.2 & 1e-3 \\
Kolkata & 128 & 64 & 0.1 & 1e-3 \\
\hline
\end{tabular}
\label{tab:hyperparams}
\end{table}

The Bayesian optimization process revealed systematic architectural preferences across the four cities. Bengaluru's model converged to a compact configuration with 32 convolutional and 16 ConvLSTM filters, utilizing 40\% dropout for regularization. Mumbai and Delhi adopted intermediate architectures with 64 convolutional and 32 ConvLSTM filters, employing 20\% dropout. Kolkata required the largest architecture with 128 convolutional and 64 ConvLSTM filters, using only 10\% dropout. A key finding was the consistent selection of 3×3 convolutional kernels across all cities. This kernel size means the model's fundamental processing unit analyzes a 3x3 cell block of the input grid, focusing on a very localized area. This uniform choice demonstrates that the models learned most effectively by identifying patterns within immediate geographical neighborhoods, prioritizing short-range spatial dependencies in the data that are characteristic of precipitation dynamics at this scale.

The ConvLSTM models displayed spatially coherent learning across India’s diverse urban rainfall regimes, but with systematic variations in performance and learned patterns aligning with each city’s dominant climate processes. In general, the Bengaluru model achieved the highest accuracy (lowest error), while Kolkata’s model had the largest errors, with Mumbai and Delhi intermediate (see Table~\ref{tab:performance}). To understand these differences, we examine a suite of post-hoc interpretability outputs for each city’s model. We organize the discussion by interpretability method—feature importance, counterfactual perturbation, Grad-CAM spatial overlays, and temporal occlusion—comparing how each model’s behavior reflects the local meteorological drivers. Throughout, we ground the interpretation in the figures for all four cities, highlighting both commonalities and contrasts across this convective-to-monsoonal spectrum.

\subsection{Feature Importance Across Cities}

\begin{figure*}[p]
\centering
\vspace{-2mm}
\resizebox{1.0\textwidth}{!}{%  % width full, height auto
\begin{tabular}{cc}
\includegraphics[width=0.47\textwidth,height=0.37\textwidth]{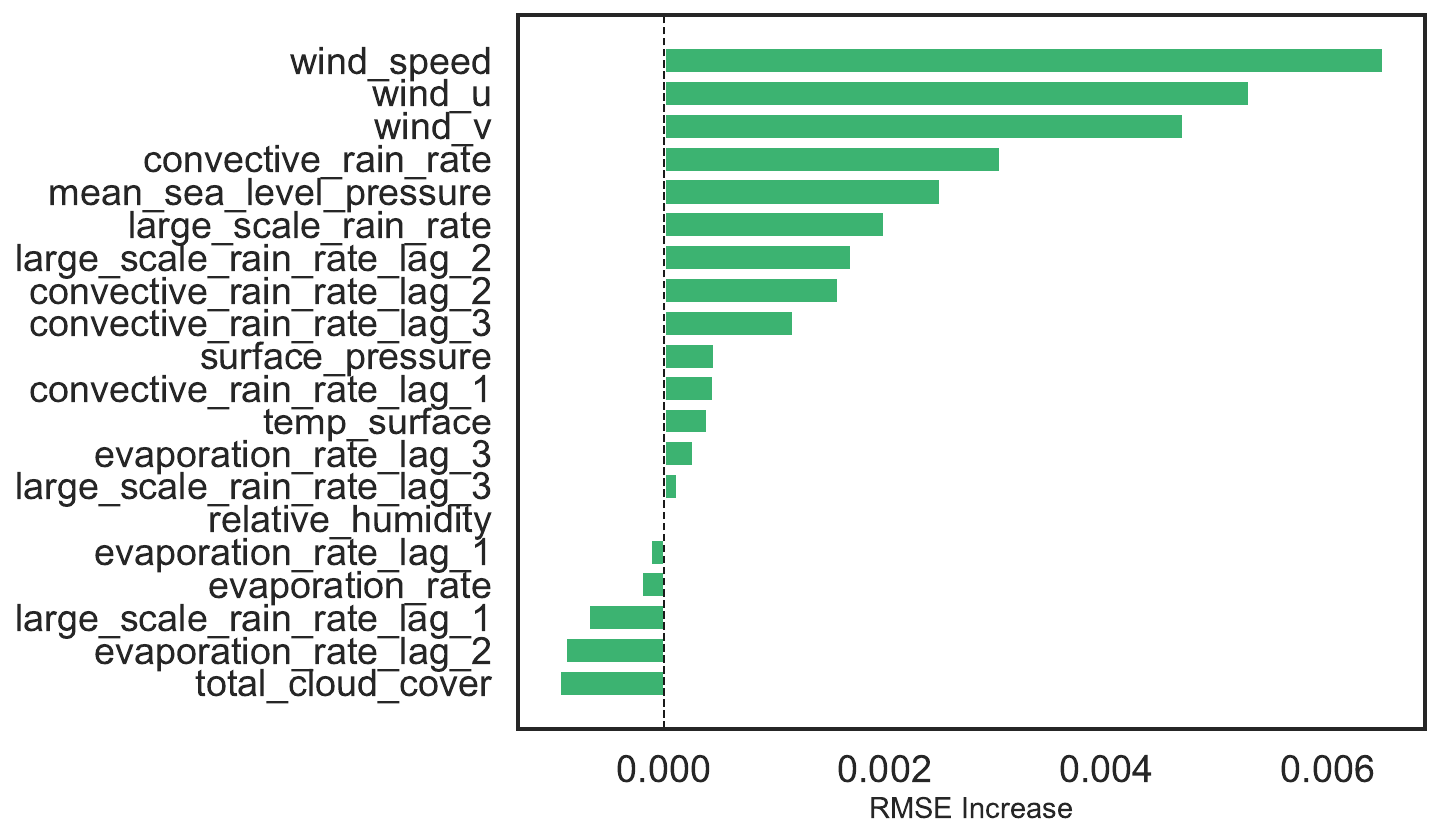} &
\includegraphics[width=0.47\textwidth,height=0.37\textwidth]{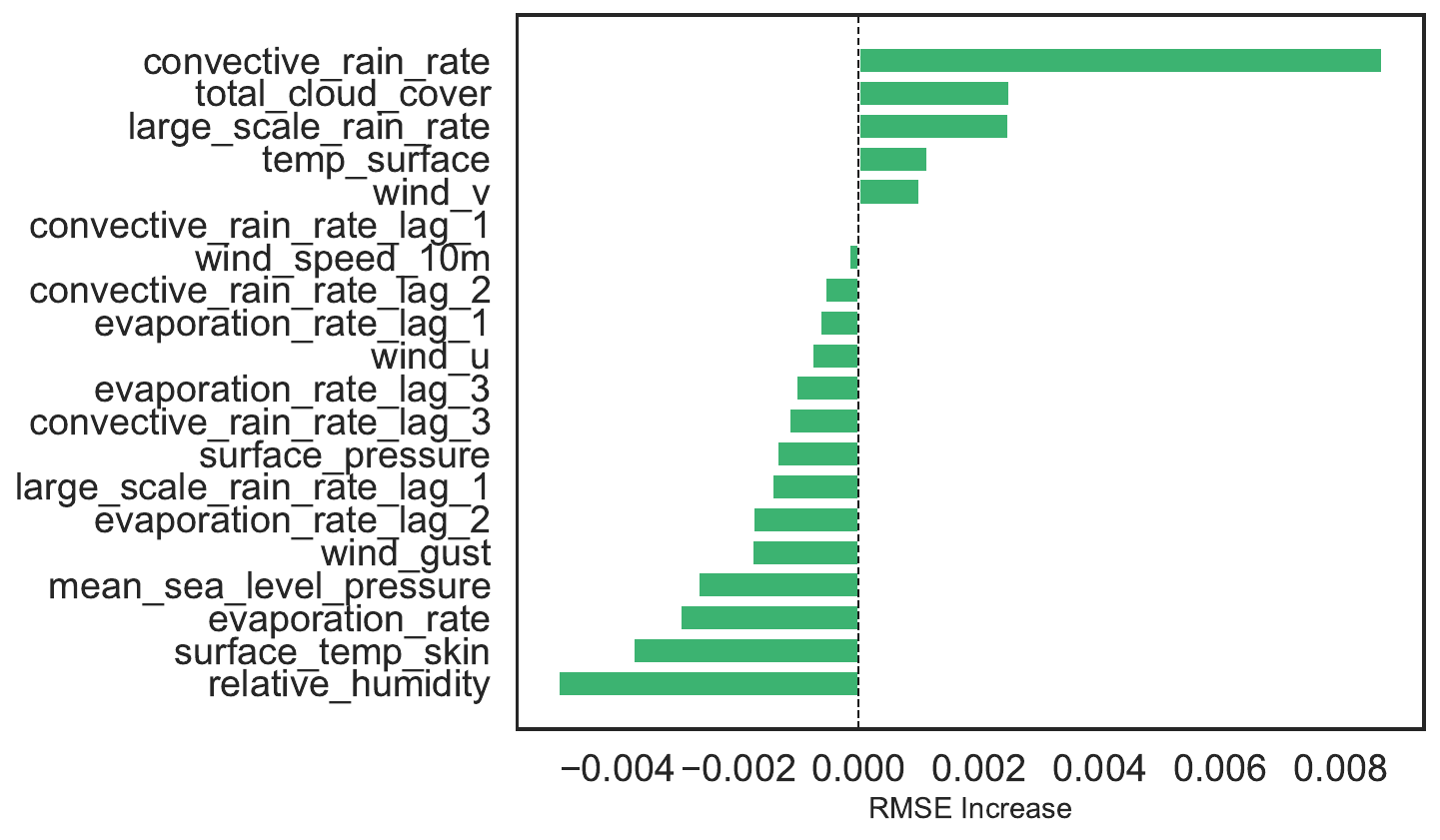} \\
\includegraphics[width=0.47\textwidth,height=0.37\textwidth]{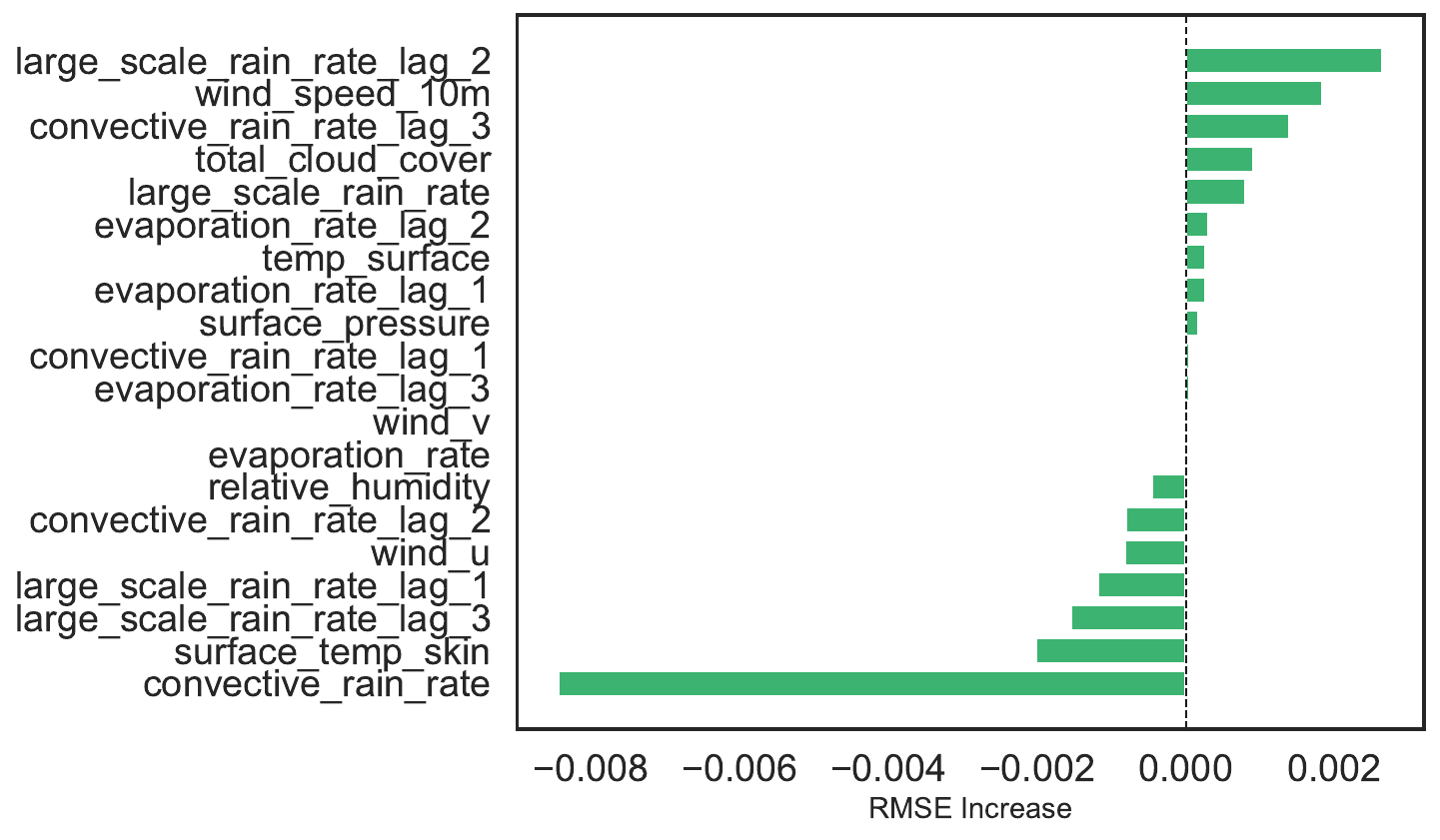} &
\includegraphics[width=0.47\textwidth,height=0.37\textwidth]{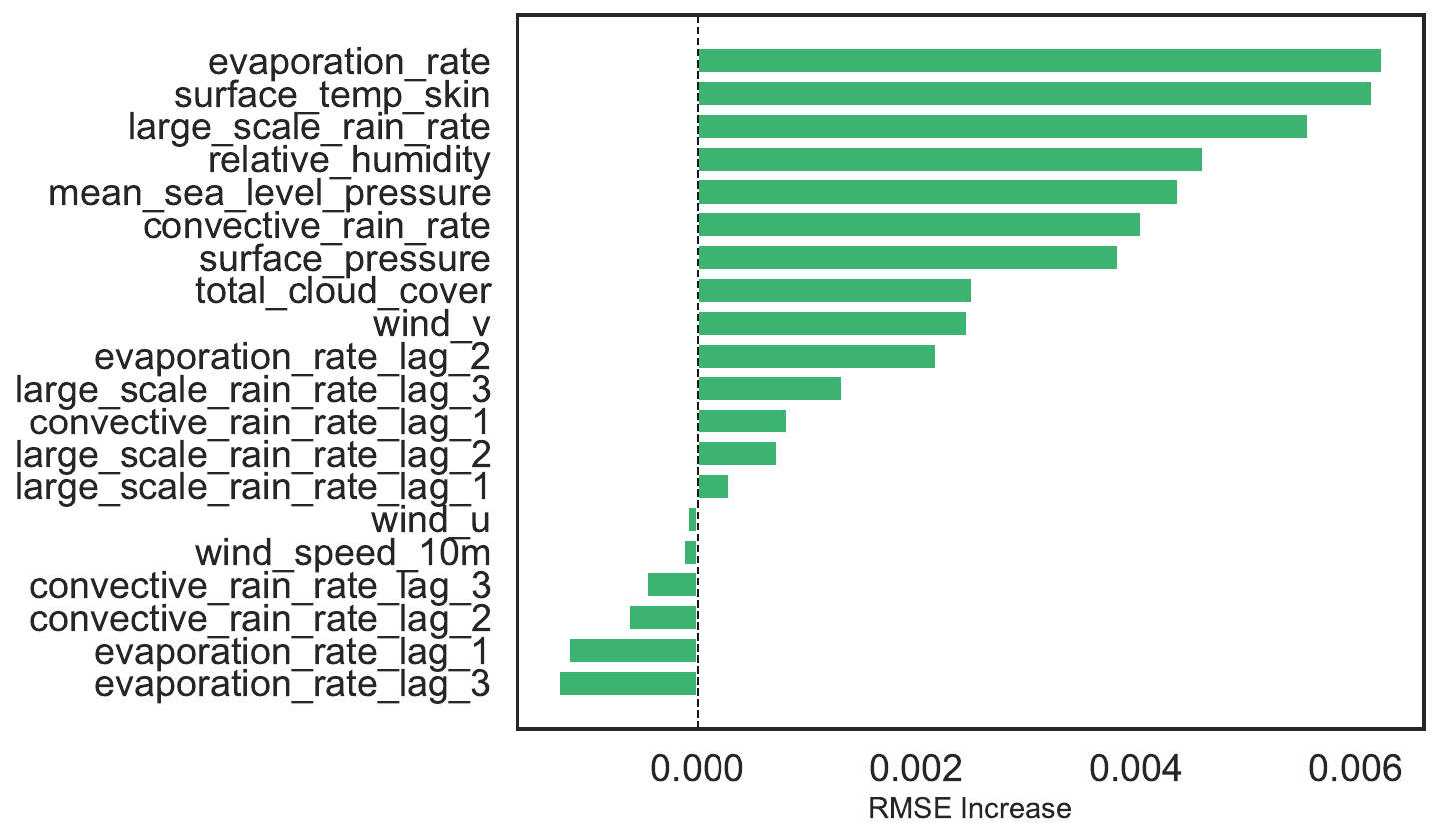}
\end{tabular}}
\vspace{-3mm}
\caption{Permutation-based feature importance for the ConvLSTM models in four cities: (a) Bengaluru, (b) Mumbai, (c) Delhi, and (d) Kolkata. The length and sign of each bar represent the change in RMSE when that input feature is randomly shuffled; positive values indicate a deterioration in performance (higher error), while values near zero or slightly negative indicate that the model can largely substitute that predictor with others.}

\label{fig:featimp}
\vspace{-3mm}
\end{figure*}

Permutation-based feature importance analysis reveals distinct key predictors for each city. The vertical axis in Figure~\ref{fig:featimp} shows the increase in root mean square error (RMSE) — expressed in millimetres per day (mm/day) — when each feature is permuted, which is in the same units as the model's precipitation output.

For Bengaluru (Figure~\ref{fig:featimp}a), the ConvLSTM model relies most strongly on wind-related variables, mean sea-level pressure, and convective  rain rate. Shuffling 10~m wind speed produces the largest decrease in model's performance, with an increase in RMSE of about 0.005–0.006 mm, followed by the zonal and meridional wind components (approximately 0.004–0.005 mm each). Randomizing mean sea-level pressure and convective rain rate increases RMSE by roughly 0.003–0.004 mm. Large-scale rain rate and recent lags of convective rain rate have a smaller, but still non-trivial, effect (about 0.002–0.003 mm). Relative humidity, evaporation rate, cloud cover, and surface temperature contribute relatively little to the Bengaluru model, with $\Delta\text{RMSE} \lesssim 0.001$ mm.

In contrast, Mumbai’s feature importance profile (Figure~\ref{fig:featimp}b) is dominated by  rain and cloud related variables. Convective rain rate (about 0.007–0.008 mm) and total cloud cover (about 0.006–0.007) produce the largest increases in RMSE when permuted, followed by large-scale rain rate (around 0.004–0.005 mm). Surface temperature and  wind components have noticeably smaller effects (approximately 0.002–0.004 mm), indicating that the Mumbai model is more tightly controlled by local convective intensity and cloudiness than by near-surface dynamics.

Delhi shows a much flatter importance distribution (Figure~\ref{fig:featimp}c). Most permutation scores lie in a narrow band around zero (roughly $-0.008$ to 0 mm), with several predictors yielding slightly negative values. , Second lag of large-scale rain rate, 10 meter wind speed and third lag of convective rain rate produced increase in RMSE. 
We interpret these near-zero and mildly negative values as evidence of interchangeability between predictors due to sampling variability, rather than as indicators of physical irrelevance: the Delhi model does not depend strongly on any single variable, and several inputs can substitute for one another.

Kolkata’s model (Figure~\ref{fig:featimp}d) stands out with a feature importance pattern weighted toward large-scale monsoonal and thermodynamic indicators. The largest RMSE increases arise from shuffling evaporation rate (about 0.005–0.006 mm), followed by skin surface temperature (around 0.004–0.005 mm) and large-scale rain rate (about 0.004). Relative humidity (approximately 0.003–0.004 mm), mean sea-level pressure, and convective rain rate exert a moderate influence (about 0.002–0.003 mm), while wind-related variables have only a small effect on model error (around 0.001–0.002 mm). Together, these results suggest a stronger role for land–atmosphere coupling and large-scale monsoon structure in Kolkata, compared with the more wind-dominated sensitivity in Bengaluru and the rain–cloud dominance in Mumbai.

\clearpage

\subsection{Counterfactual perturbation}

\begin{figure}[H]
\centering
\vspace{-2mm}
\resizebox{1.15\textwidth}{!}{%
  \begin{tabular}{cc}
    \includegraphics[width=0.49\textwidth,height=0.40\textwidth]{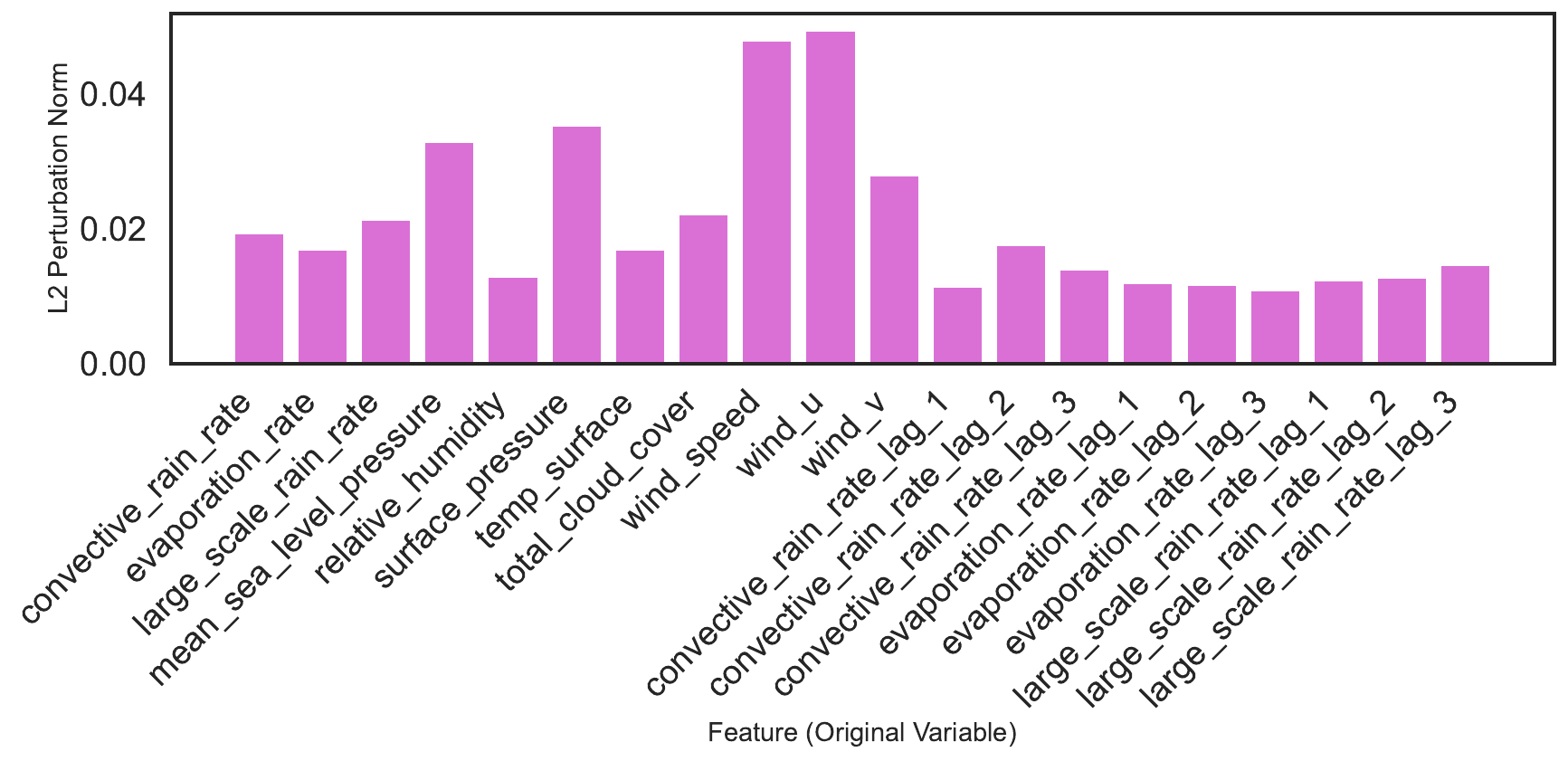} &
    \includegraphics[width=0.49\textwidth,height=0.40\textwidth]{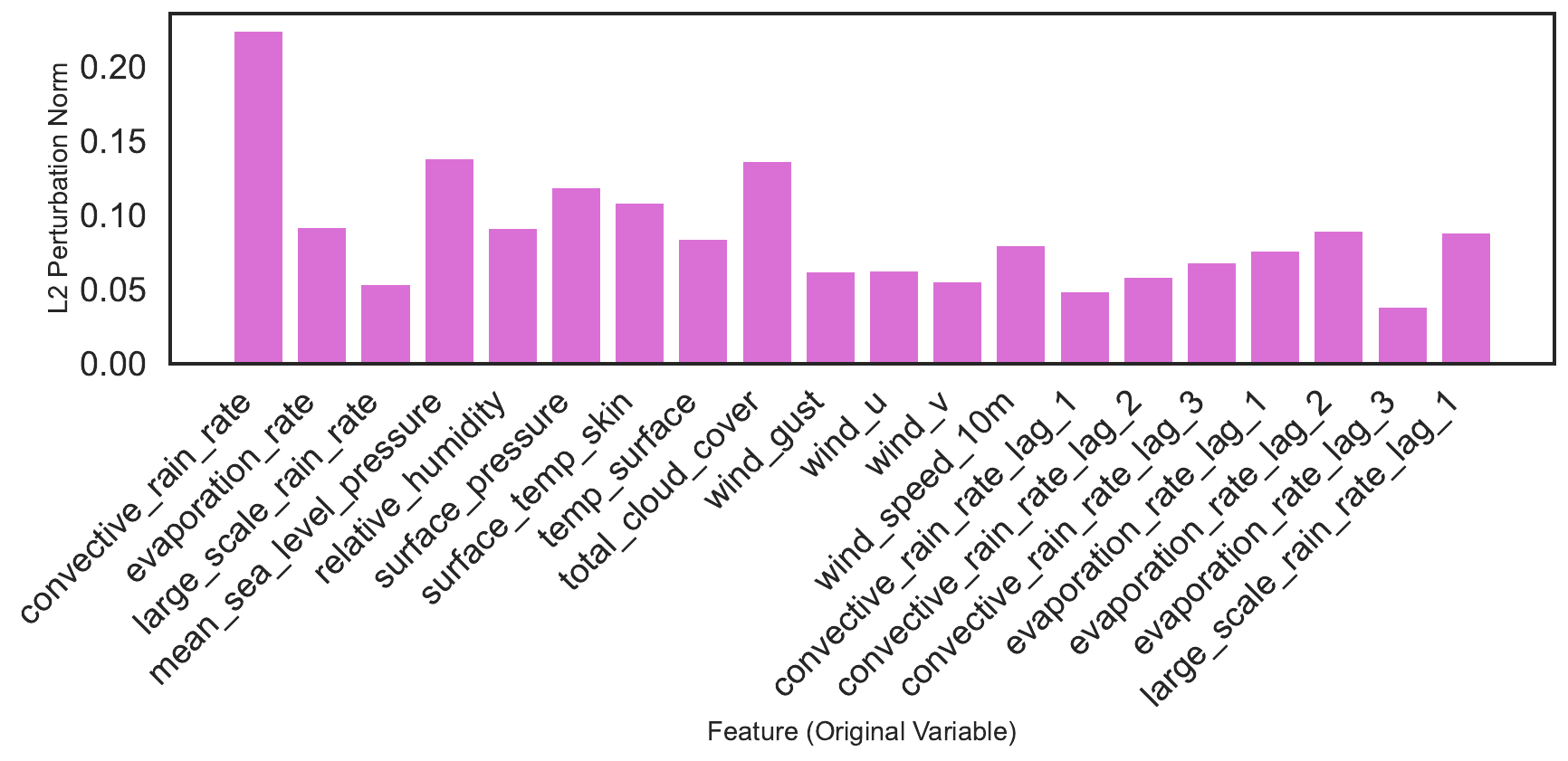} \\
    \includegraphics[width=0.49\textwidth,height=0.40\textwidth]{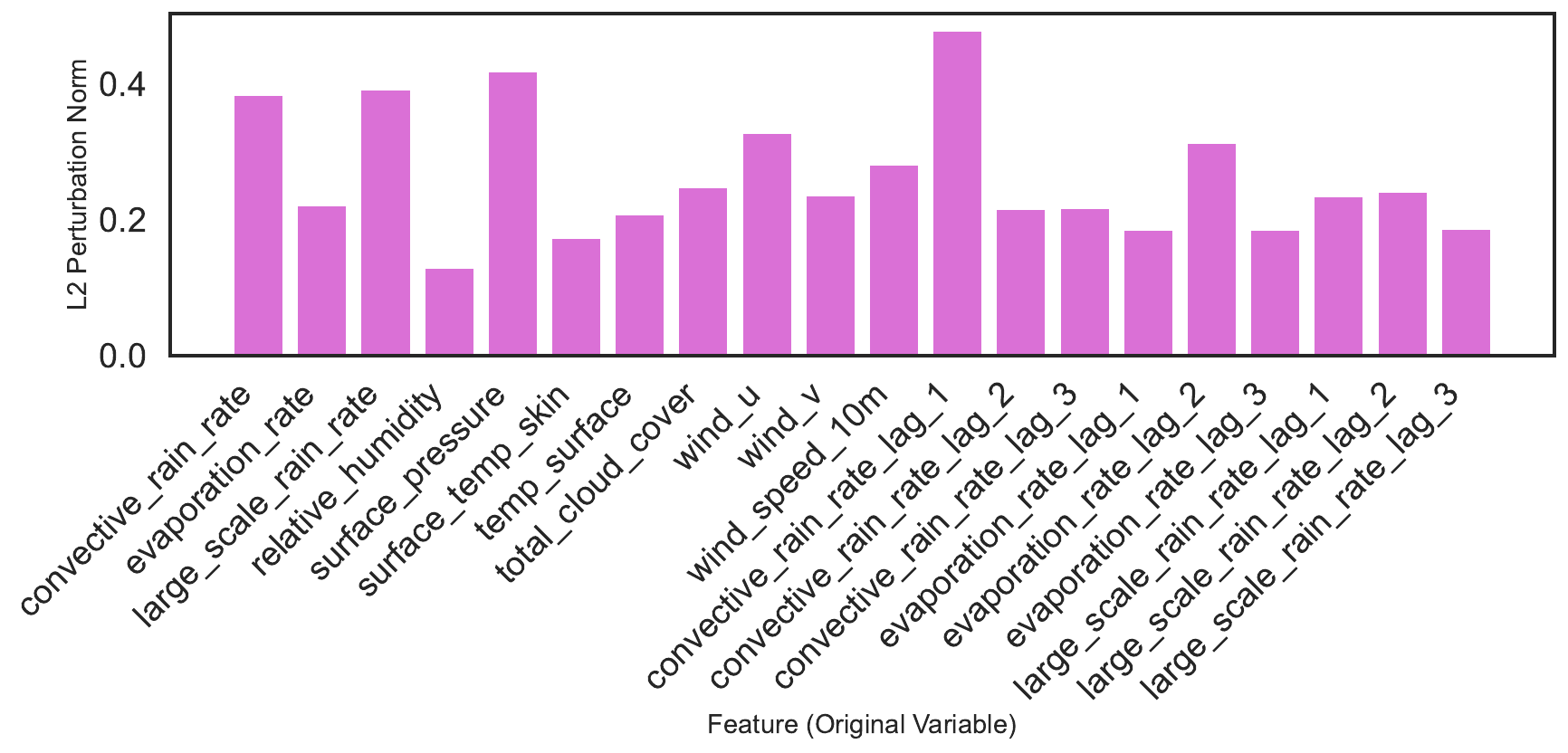} &
    \includegraphics[width=0.49\textwidth,height=0.40\textwidth]{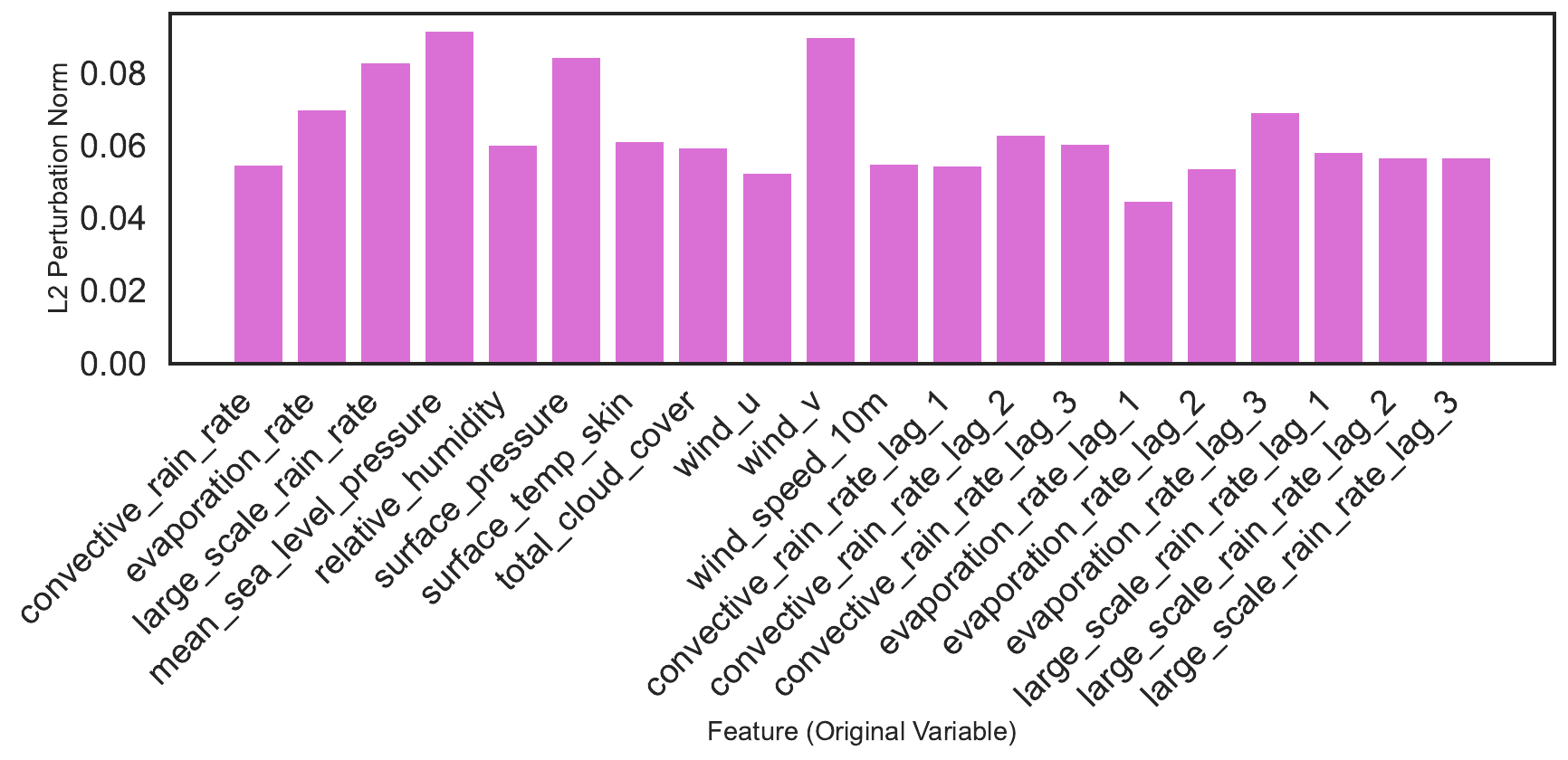}
  \end{tabular}%
}
\vspace{-2mm}
\caption{Counterfactual perturbation analysis for the ConvLSTM models in four cities: (a) Bengaluru, (b) Mumbai, (c) Delhi, and (d) Kolkata. Each panel shows how predicted rainfall changes when individual input features are systematically reduced, indicating model sensitivity to each meteorological variable.}
\label{fig:counterfactual}
\vspace{-3mm}
\end{figure}

The counterfactual perturbation analysis examines how much the forecast changes when each input variable is slightly reduced, while all other inputs are kept fixed. The vertical axis in Figure~\ref{fig:counterfactual} reports the L2 perturbation norm of the change in predicted rainfall, expressed in millimetres (mm), that is, in the same units as the model output.

For Bengaluru (Figure~\ref{fig:counterfactual}a, the largest responses occur for the wind variables: the zonal and meridional wind components show the highest perturbation norms (around 0.04–0.045\,mm), followed by wind speed (around 0.03\,mm). Mean sea-level pressure and surface pressure form the next group, with smaller but still noticeable changes, while rainfall, cloud and humidity variables have comparatively lower values.

For Mumbai (Figure~\ref{fig:counterfactual}b), the model is most sensitive to convective rain rate, which produces by far the largest perturbation norm (about 0.20–0.22\,mm). A second tier of variables, including evaporation rate, relative humidity, surface pressure, skin surface temperature and total cloud cover, shows moderate changes. The remaining variables, including wind components and lagged rainfall terms, have smaller but non-negligible effects.

For Delhi (Figure~\ref{fig:counterfactual}c), the largest perturbation norm is associated with convective rain rate (lag 2), with a value close to 0.45\,mm. The current convective rain rate and large-scale rain rate also show strong responses (around 0.35–0.40\,mm). Evaporation rate and several other moisture-related variables form a middle group, while many other predictors, including the wind variables, produce more moderate changes in the forecast when they are slightly reduced.

For Kolkata (Figure~\ref{fig:counterfactual}d), the largest perturbation norms lie in a narrow range between about 0.07 and 0.09\,mm and are shared by several variables, including mean sea-level pressure, evaporation rate, relative humidity, one of the wind components and large-scale rain rate. Skin surface temperature and the remaining inputs produce smaller, but still measurable, changes. Overall, the Kolkata model appears to rely on a broader cluster of variables rather than a single dominant one.

\subsection{Spatial attention using Grad-CAM}

Grad-CAM activation maps, overlaid with city boundaries, show which geographical regions the ConvLSTM model focuses on when producing precipitation forecasts. These maps therefore provide spatial context for the model's learned decision strategy, but they should be interpreted as descriptions of model behaviour, not as proofs of physical mechanisms.

\begin{figure}[H]
\centering
% --- First row ---
\includegraphics[width=0.48\textwidth,trim=10 40 10 20,clip]{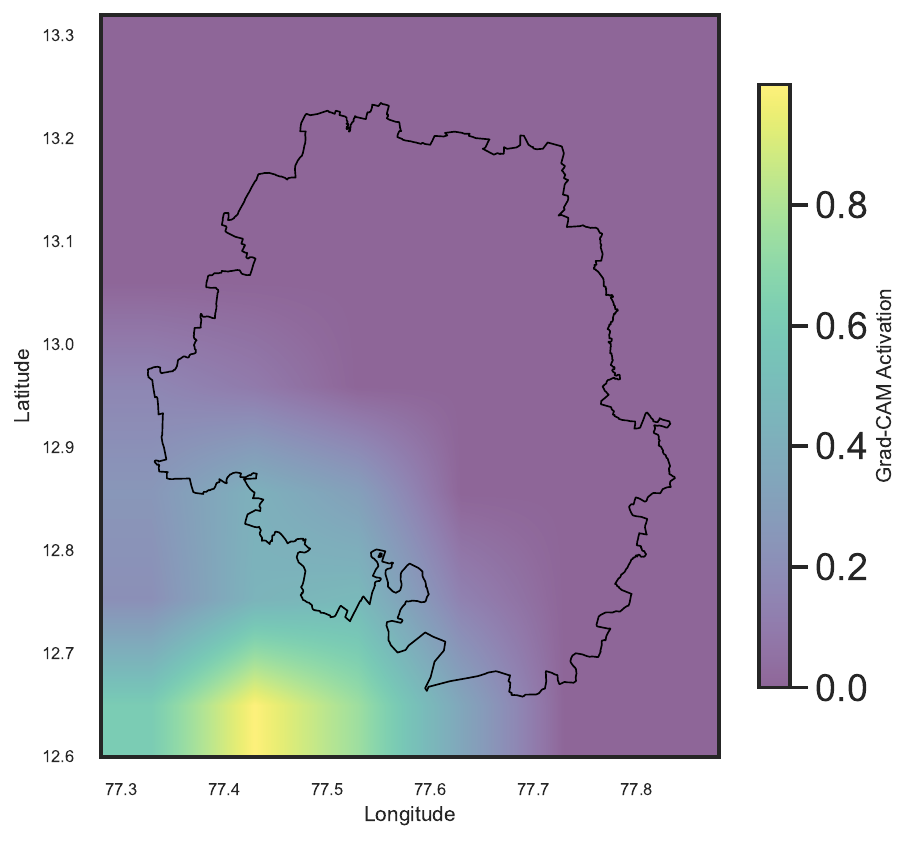}
\includegraphics[width=0.48\textwidth,trim=10 110 10 40,clip]{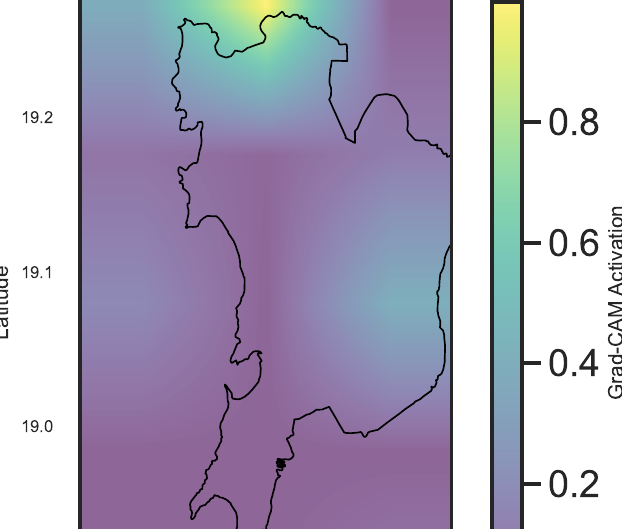}\\[1ex]
% --- Second row ---
\includegraphics[width=0.48\textwidth,trim=10 40 10 20,clip]{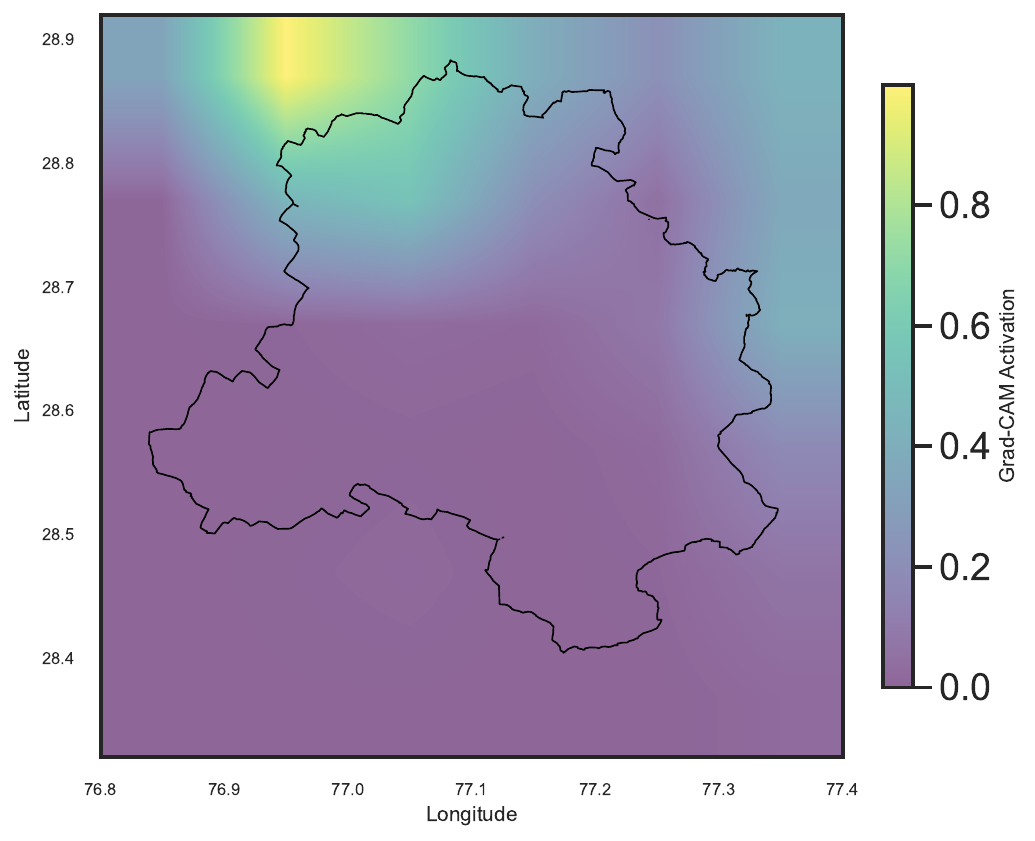}
\includegraphics[width=0.48\textwidth,trim=10 40 10 20,clip]{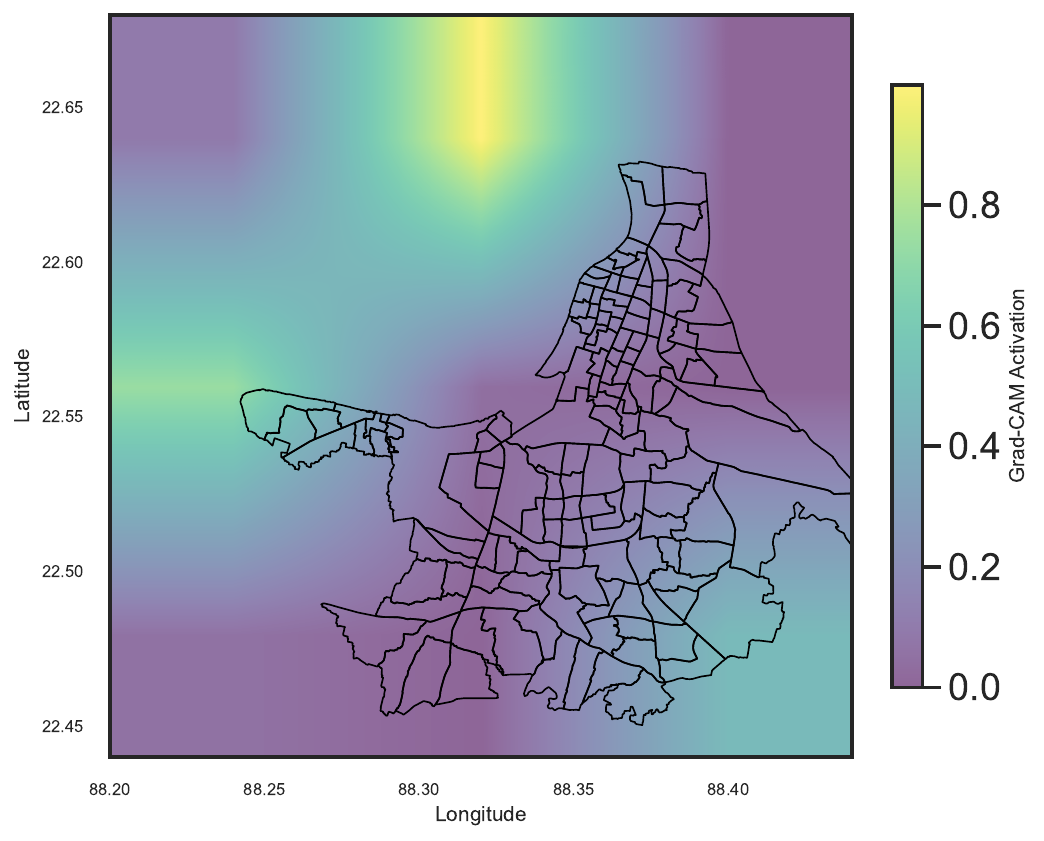}
\caption{Grad-CAM activation overlays showing representative high-rain predictions for (a)~Bengaluru, (b)~Mumbai, (c)~Delhi, and (d)~Kolkata. Warmer colors denote regions where the ConvLSTM model is paying the most attention (i.e., contributing strongly to the predicted rainfall) at forecast time.}
\label{fig:gradcam}
\end{figure}

Bengaluru's Grad-CAM overlay (Figure~\ref{fig:gradcam}a) shows a compact region of strongest activation over the southwestern part of the domain, near the city's hilly periphery, with weaker values elsewhere. This pattern indicates that, for high-rain forecasts, the model consistently pays more attention to that part of the input grid, which is consistent with the presence of known convective activity in this sector, although we do not attempt a full physical explanation here.

Mumbai's Grad-CAM overlay (Figure~\ref{fig:gradcam}b) shows comparatively stronger activation along the northern part of the city close to the Arabian Sea coastline and adjacent near-shore area, with weaker responses over the southern and more inland sections. This suggests that the model relies more heavily on conditions in the northern coastal belt when producing high-rain forecasts for Mumbai.

Delhi's Grad-CAM overlay (Figure~\ref{fig:gradcam}c) shows a localized area of stronger activation in the northern part of the domain, with comparatively weaker responses over central and southern Delhi. This pattern suggests that the model is more sensitive to input conditions over the northern portion of the domain when issuing high-rain forecasts for the city.

Kolkata's Grad-CAM overlay (Figure~\ref{fig:gradcam}d) shows stronger activation over the northern and north-eastern parts of the domain, with weaker responses over the central and southern portions of the city. This indicates that, for high-rain forecasts in Kolkata, the model draws more heavily on information from these northern and north-eastern grid cells.

\subsection{Temporal Occlusion (Memory) Analysis}

\begin{figure}[H]
\centering
\vspace{-2mm}
\resizebox{1.15\textwidth}{!}{%
  \begin{tabular}{cc}
    \includegraphics[width=0.49\textwidth,height=0.40\textwidth]{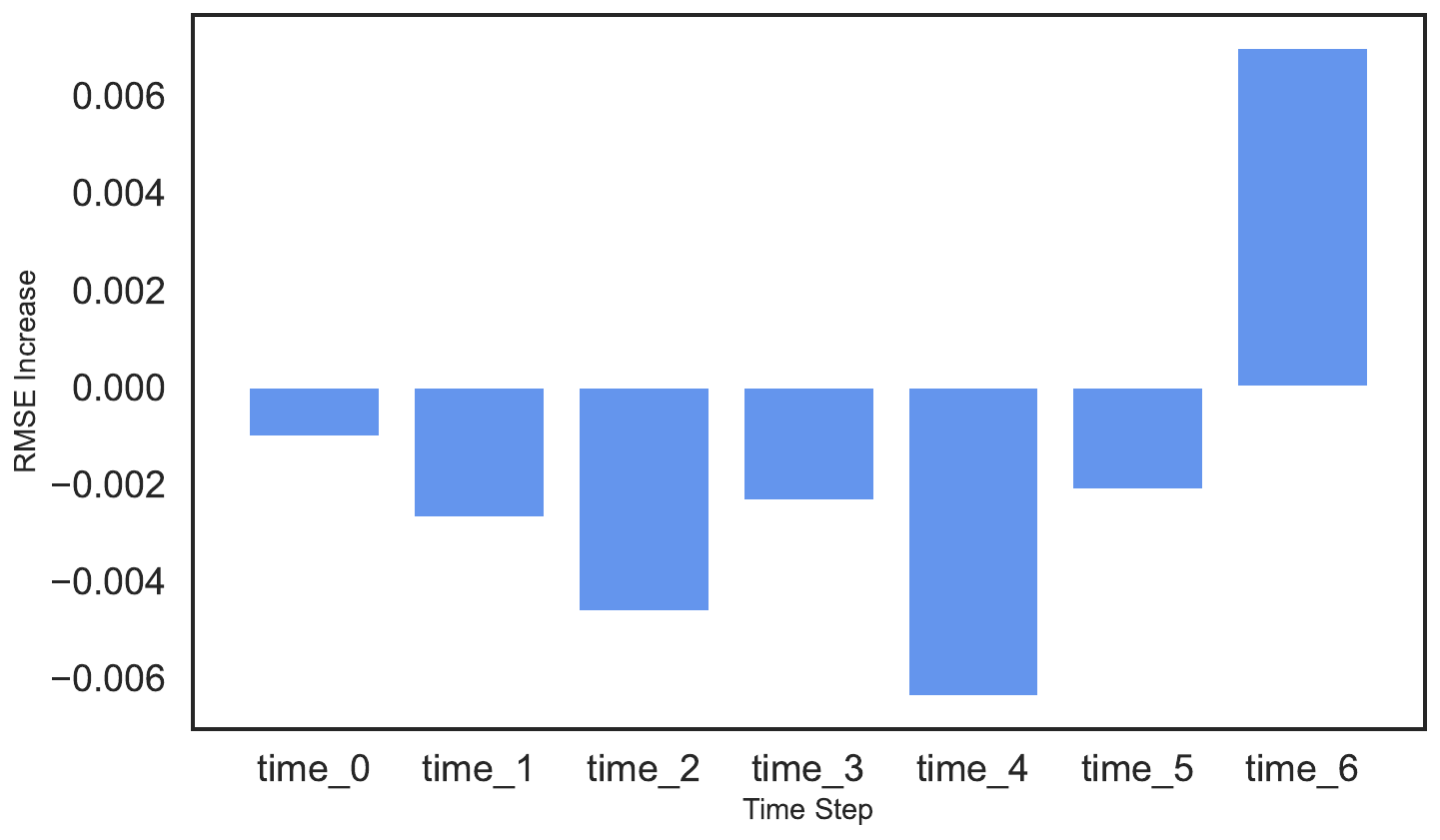} &
    \includegraphics[width=0.49\textwidth,height=0.40\textwidth]{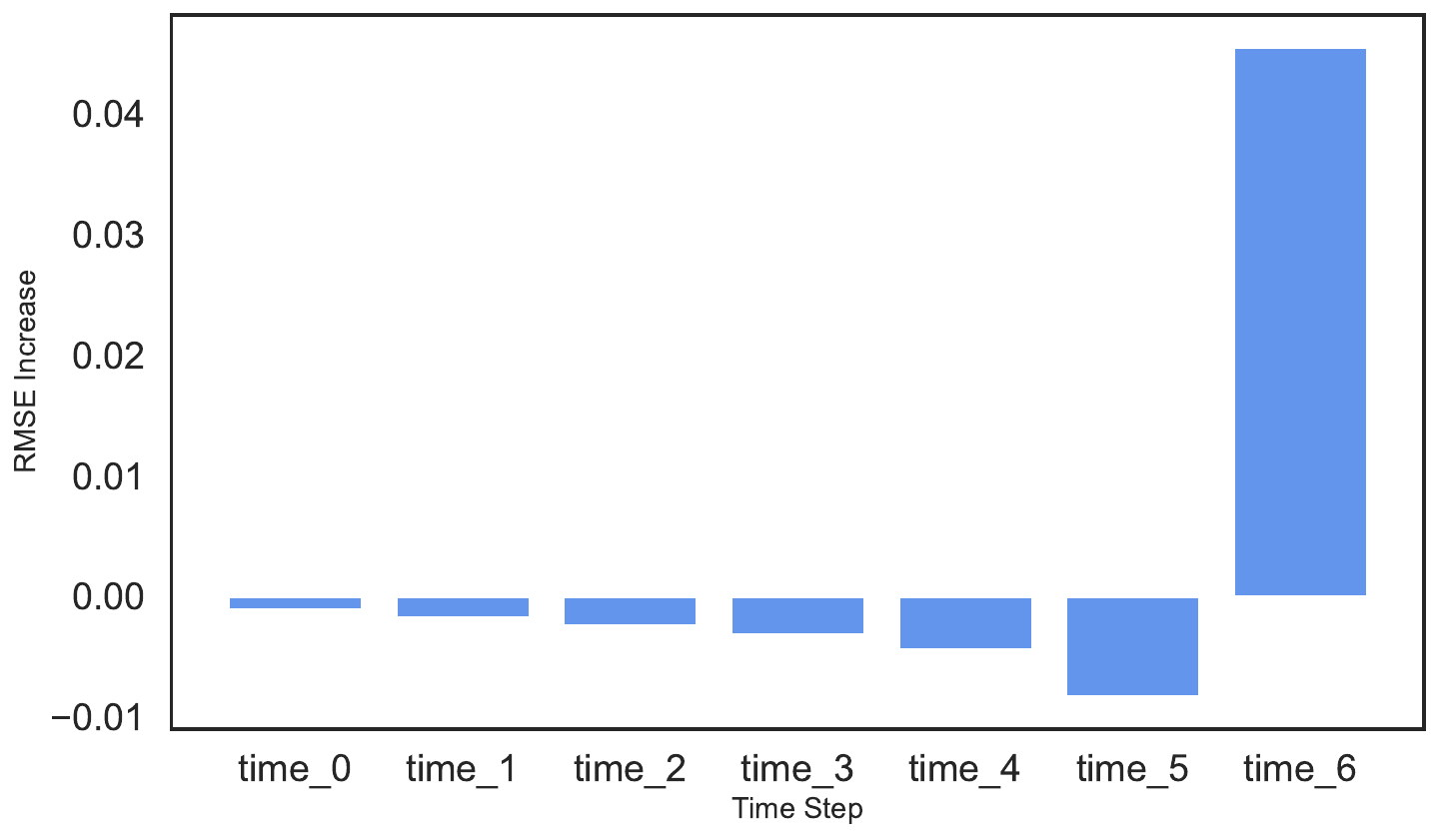} \\
    \includegraphics[width=0.49\textwidth,height=0.40\textwidth]{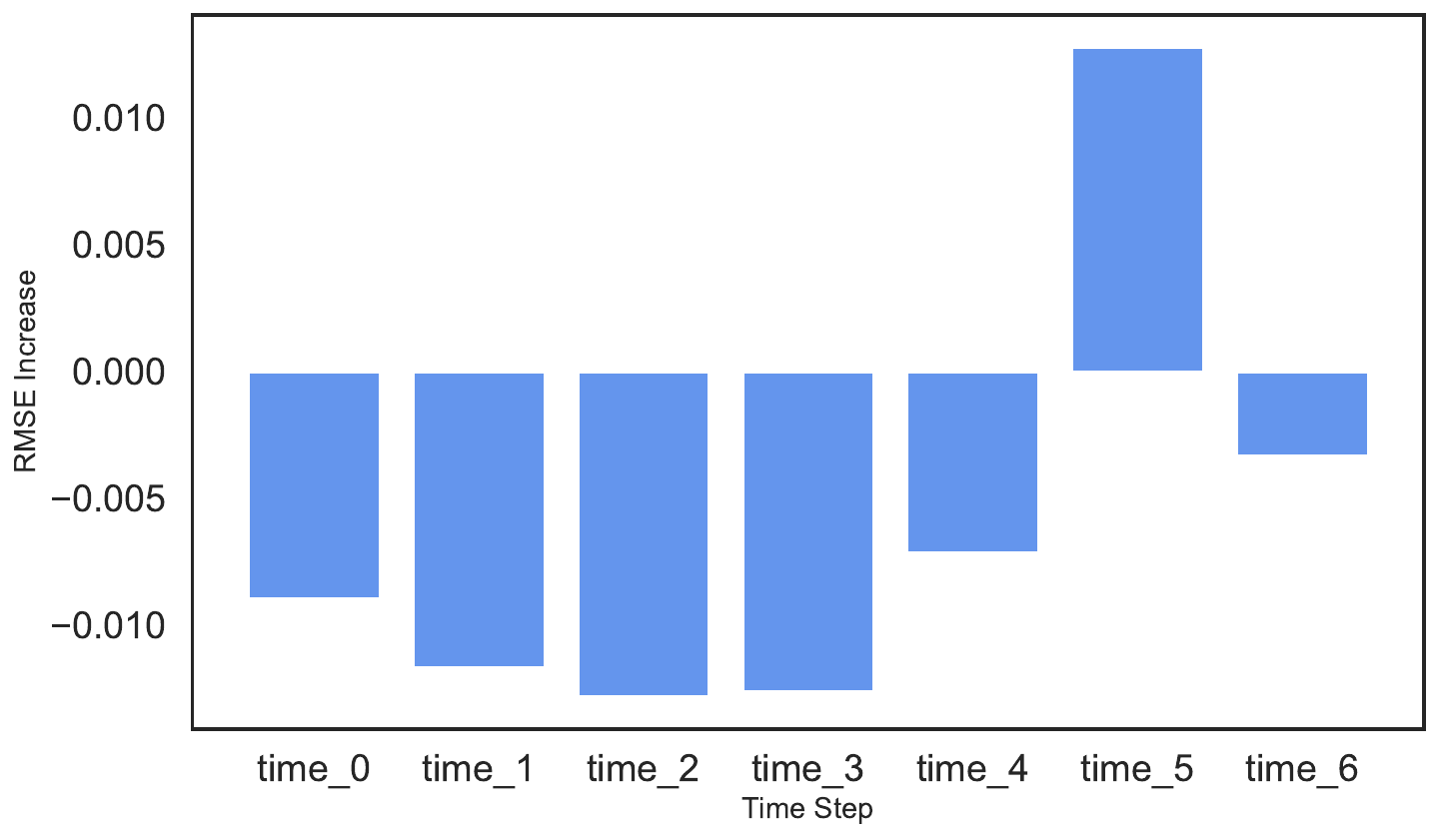} &
    \includegraphics[width=0.49\textwidth,height=0.40\textwidth]{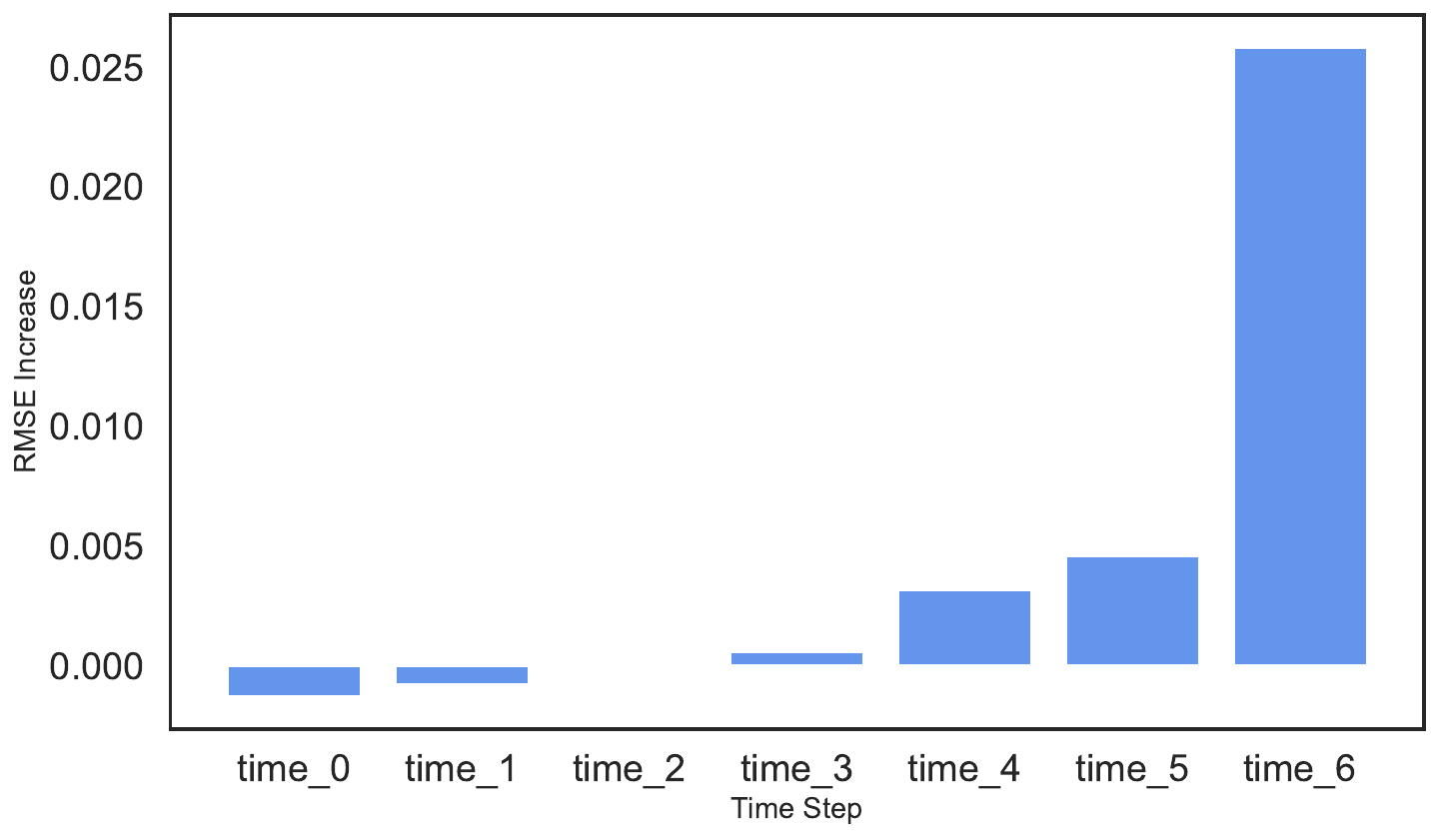}
  \end{tabular}%
}
\vspace{-2mm}
\caption{Occlusion-based time-step importance for the ConvLSTM models in four cities: (a) Bengaluru, (b) Mumbai, (c) Delhi, and (d) Kolkata. Each panel shows the increase in prediction error ($\Delta$RMSE) when data from a given past day is masked, indicating how much each lag contributes to the forecast.}
\label{fig:occlusion}
\vspace{-3mm}
\end{figure}

We examine the temporal dependencies of each city’s ConvLSTM model using an occlusion analysis of the input time steps. Figure~\ref{fig:occlusion} shows how the prediction error changes when data from each day in the 7-day input sequence (time\_0 to time\_6, with time\_6 denoting the most recent day) are masked in turn. Larger positive values indicate that removing that day’s information declines the forecast more strongly, while values close to zero or slightly negative suggest that the model can largely substitute information from other days.
For Bengaluru (Figure~\ref{fig:occlusion}a), masking the most recent day (time\_6) produces the only clear increase in error, with $\Delta\text{RMSE}$ on the order of 0.006–0.007 mm. Occluding earlier days (time\_0–time\_5) yields changes that are small and sometimes slightly negative. This pattern indicates that the Bengaluru model relies mainly on very recent conditions, with limited sensitivity to information from earlier days in the input window.
Mumbai’s model (Figure~\ref{fig:occlusion}b) shows an even more pronounced dominance of the last input day: occluding time\_6 increases RMSE by roughly 0.04–0.045 mm, whereas all other time steps produce changes close to zero and one (time\_5) is slightly negative. In practice, this means that the Mumbai forecasts are strongly anchored in the most recent day’s fields, and the remaining lags contribute relatively little additional skill in this occlusion experiment.
Delhi’s occlusion profile (Figure~\ref{fig:occlusion}c) is different. Here, the largest positive impact appears when masking time\_5, with $\Delta\text{RMSE}$ of about 0.01–0.012 mm, while the other time steps, including time\_6, are near zero or slightly negative. This suggests that, for Delhi, the model is somewhat more sensitive to conditions one step back in time than to the very last input day, although overall the magnitudes remain modest.
Kolkata (Figure~\ref{fig:occlusion}d) shows a more gradually increasing pattern across the most recent lags. Occluding time\_3 and time\_4 produces small but positive increases in RMSE, time\_5 gives a larger change, and masking the most recent day (time\_6) yields the largest increase, on the order of 0.02–0.025 mm. This indicates that several of the last few days carry useful information for the Kolkata model, with a clear peak in importance for the most recent conditions.

\section{Discussion and conclusion}

This study demonstrates that a ConvLSTM model can generate accurate one-day-ahead rainfall forecasts for four Indian megacities while providing transparency into its decision-making process. The model effectively captured daily rainfall variability, including high-intensity events, with predictive performance varying across climatic regimes: errors were lowest for Bengaluru and Mumbai, moderate for Delhi, and highest for Kolkata. This indicates the architecture's adaptability to diverse settings, despite inherent differences in precipitation predictability.

A suite of interpretability techniques was employed to elucidate the physical consistency of the model's predictions. Two complementary experiments provided particular insight into variable importance. The permutation importance test quantifies a variable's overall contribution by measuring the increase in RMSE when its values are randomly permuted; a large increase indicates the model relies heavily on that input. In contrast, counterfactual perturbation analysis assesses local sensitivity by measuring the change in prediction (L$_2$ norm) when a variable is systematically reduced. Together, these methods identify inputs that are both critical to overall performance and those to which predictions are locally sensitive.

The integrated interpretability analysis reveals distinct city-specific reliance patterns. For Bengaluru, the three wind variables are unequivocally dominant, exhibiting the largest permutation importance and counterfactual changes, with a secondary group of convective rain rate, large scale rain rate, and mean sea level pressure showing medium to high importance in both tests, while cloud and humidity variables are consistently less influential. For Mumbai, convective rain rate has the largest permutation importance and counterfactual change, with large scale rain rate also ranking highly in both; however, total cloud cover is important when permuted but only mid-range in counterfactual sensitivity, while evaporation rate and relative humidity show the opposite pattern, and wind variables are relatively unimportant. For Delhi, permutation importance is universally small and mixed, with variables like large scale rain rate at lag 2 and convective rain rate at lag 3 showing weak positive increases, indicating that predictors are largely substitutable when randomized; conversely, the counterfactual test shows sharp local sensitivity to a subset of moisture and rain related variables. Finally, for Kolkata, permutation importance identifies evaporation rate, skin surface temperature, large scale rain rate, and relative humidity as the top group, followed by mean sea level pressure; the counterfactual plot shows evaporation rate and large scale rain rate again in the leading cluster, joined by the meridional wind component, with relative humidity and skin surface temperature remaining highly influential, indicating consistent importance for evaporation and large scale rain with strong but not exclusive roles for humidity and skin temperature and an additional wind influence evident primarily in the counterfactual experiment.
The temporal analysis reveals distinct memory dependencies across cities: Bengaluru and Mumbai models rely predominantly on the most recent day's data, whereas Delhi's model shows greater dependence on the day-before-last, and Kolkata's utilizes a multi-day sequence with increasing importance up to the present. Spatially, Grad-CAM visualizations demonstrate that each model consistently focuses on specific sub-regions when predicting precipitation—Bengaluru on a compact southwestern zone, Mumbai on a narrow northern coastal strip, Delhi on a broad northern band extending southeast, and Kolkata on a wide northeastern area with a secondary eastern ridge. These spatiotemporal patterns, combined with the variable-wise importance, provide a complete interpretable framework clarifying which inputs, from where, and from when most strongly influence each city's rainfall forecasts.

To contextualise these model-derived patterns, we compare them with findings from previous scientific studies about rainfall in each city. For Kolkata, earlier research has shown that rainfall is heavily influenced by atmospheric moisture and stability, which affects flood risk \citep{Dasgupta2013,De2022}. The model's focus on evaporation rate, skin surface temperature, relative humidity, and large-scale rain rate, along with its use of a multi-day memory, matches this established understanding of how prolonged moist conditions lead to rain. For Mumbai, previous analyses have connected extreme rainfall to large-scale wind patterns and urban effects \citep{Chalakkal2022}. The model's strong dependence on convective rain rate and large-scale rain rate, together with its narrow spatial focus on the coastline, is consistent with the known importance of ocean-influenced weather systems for heavy rain in this region. In Delhi, studies have noted that rainfall extremes are linked to broad regional features \citep{Rana2012}. The model's behaviour here---where no single variable is critically important, but its predictions are sensitive to reductions in moisture, and it pays attention to a wide area---fits the concept that Delhi's rain comes from a combination of many regional factors rather than one local cause. For Bengaluru, recent work has shown that brief, intense storms are a main cause of flash floods \citep{Bindal2022}. The model's good performance, its reliance on wind speed, zonal wind, meridional wind, and mean sea-level pressure, and its highly localised spatial focus suggest it is using the kinds of quick-changing, local weather patterns known to produce such storms.

On a national scale, other research has documented changes in short-duration and urban rainfall extremes across India, including multi-city analyses that cover Mumbai and other major urban areas \citep{Ali2014,Goswami2006,Fowler2021,Guhathakurta2011}. Other reviews emphasise that city growth and inadequate drainage systems make the impacts of these heavy rains worse \citep{Dhiman2019,Dharmarathne2024,Deopa2024}. While our study does not investigate these trends or the interaction between extreme rainfall and urban infrastructure directly, it adds a complementary viewpoint by showing that the deep learning model's behaviour in these four cities can be interpreted alongside their known and well-studied rainfall climatologies.

Several limitations should be considered when interpreting these findings. The models were trained on ERA5 reanalysis data at $0.25^{\circ}$ resolution and predict a single area-averaged rainfall value for each city. This approach cannot resolve neighbourhood-scale rainfall gradients or directly predict local flood depths. The input data are limited to meteorological variables and do not include other influential factors such as land use characteristics, antecedent soil moisture conditions, tidal patterns, or drainage capacity, all of which are known to affect rainfall impacts.

The interpretability methods themselves have certain assumptions. Permutation analysis can be affected by correlations between inputs, small counterfactual changes may move the model into rarely observed parts of input space, and Grad-CAM maps depend on network architecture and training details, as noted in previous studies \citep{GonzalezAbad2023,Wu2023,Xiang2024}. For these reasons, the results should be understood as a structured description of model behaviour rather than a complete reconstruction of atmospheric processes.

Despite these limitations, the findings offer practical value. They identify which variables in each city should be monitored most closely for data quality. The temporal analysis indicates the necessary lead time for collecting reliable input data, while the spatial attention maps show where enhanced monitoring networks would be most beneficial. Future work could build on this foundation by coupling similar ConvLSTM models with higher-resolution rainfall products and urban hydrological models, following recent flood-forecasting frameworks \citep{Davis2022}. Additional experiments could directly test the importance of key variables by measuring how forecast skill changes when these variables are systematically included or excluded.

In conclusion, this study demonstrates that ConvLSTM models can provide accurate short-range rainfall forecasts while offering transparent insights into their decision-making processes across diverse urban Indian climates. The city-specific architectures achieved predictive performance aligned with regional precipitation characteristics, from Bengaluru's localised storms to Kolkata's monsoonal systems. More significantly, the comprehensive interpretability framework revealed physically consistent patterns—identifying dominant meteorological drivers, climate-appropriate memory horizons, and geographically meaningful attention regions for each city. These model-derived patterns show descriptive compatibility with established observational studies, reinforcing their physical plausibility without claiming causation. From an operational standpoint, the findings enable more efficient resource management. They provide a clear rationale for focusing data quality efforts on a shortlist of critical variables, for defining the necessary historical data window, and for prioritising investments in enhanced observation networks within the most sensitive sub-regions around each city. By maintaining focus on explicating model behaviour rather than asserting mechanistic insight, this work establishes a foundation for developing trustworthy AI tools in urban hydrometeorology that balance predictive performance with transparent decision-making.

\bibliography{bibliography}

\end{document}